\documentclass[final,5p,times,twocolumn,authoryear]{elsarticle}
\usepackage[utf8]{inputenc}
\usepackage[T1]{fontenc}
\usepackage{amsmath,amssymb,amsthm}
\usepackage{graphicx}
\usepackage{booktabs}
\usepackage{multirow}
\usepackage{subcaption}
\usepackage{algorithm}
\usepackage{algpseudocode}
\usepackage[table]{xcolor}
\definecolor{rowshade}{gray}{0.92}
\usepackage{pgfplots}
\pgfplotsset{compat=1.17}
\usepackage{hyperref}
\usepackage{tabularx}

\newtheorem{proposition}{Proposition}

\newtheorem{observation}{Observation}

\newcommand{\sd}[1]{{\scriptsize${}\pm{}$#1}} % in the preamble

\newcommand{\inst}{x}                       % instance
\newcommand{\nfeat}[1]{x_{#1}}              % node features
\newcommand{\Nodes}{V}
\newcommand{\cost}[2]{d_{#1#2}}             % pairwise cost
\newcommand{\dem}[1]{q_{#1}}                % demand
\providecommand{\Cap}{}\renewcommand{\Cap}{Q}  % amssymb defines \Cap (unused symbol); override it
\newcommand{\sol}{\sigma}
\newcommand{\Perm}{\Pi_n}
\newcommand{\nroutes}{M}                    % routes in giant tour
\newcommand{\Len}{L}                        % solution length = #tokens
\newcommand{\Cost}{C}                       % objective C(sigma|x)
\newcommand{\edge}[1]{e_{#1}}
\newcommand{\Edges}{E}
\newcommand{\route}{r}
\newcommand{\pload}[1]{P_{#1}}              % prefix load
\newcommand{\rload}[1]{\Lambda_{#1}}        % route load
\newcommand{\move}[2]{\mu_{#1#2}}           % 2-opt move
\newcommand{\dcost}[2]{\Delta_{#1#2}}       % cost change
\newcommand{\Adm}{\mathcal{A}}
\newcommand{\Budget}{T}
\newcommand{\hist}{c}                       % search history c_t
\newcommand{\Bhist}{B_{\mathrm{hist}}}
\newcommand{\mtabu}{m_{\mathrm{tabu}}}
\newcommand{\tok}[1]{\tau_{#1}}             % ordered token
\newcommand{\elen}[1]{d_{#1}}               % edge length
\newcommand{\rellen}[1]{\mathrm{rel\_len}_{#1}}
\newcommand{\zs}[1]{\zeta_{#1}}             % length z-score
\newcommand{\fhist}[1]{f^{\mathrm{hist}}_{#1}}
\newcommand{\knnlen}[1]{\lambda_{#1}}      % length / local k-NN spacing
\newcommand{\spacing}[1]{\nu_{#1}}        % mean distance to the 10 nearest nodes
\newcommand{\emb}[1]{h_{#1}}
\newcommand{\cemb}[1]{\tilde h_{#1}}
\newcommand{\gsum}{\bar h}             % global summary (mean token)
\newcommand{\dmodel}{D}
\newcommand{\Nblk}{N_{\mathrm{blk}}}
\newcommand{\dkey}{d_{\mathrm{key}}}
\newcommand{\cmix}{\gamma}                  % cycle-mixing scale
\newcommand{\scorer}{\psi}

\newcommand{\logit}[2]{\ell_{#1#2}}
\newcommand{\slogit}[2]{\bar\ell_{#1#2}}
\newcommand{\lmax}{c_{\max}}
\newcommand{\gain}[2]{\omega_{#1#2}}      % normalized exact gain of a move
\newcommand{\gbeta}{\beta}                % learned gain weight
\newcommand{\gmax}{\omega_{\max}}
\newcommand{\pol}{p_\theta}
\newcommand{\polref}{p_{\theta_{\mathrm{ref}}}}
\newcommand{\guard}{\eta}
\newcommand{\Dist}{\mathcal{D}}
\newcommand{\nlow}{n_{\mathrm{low}}}
\newcommand{\nhigh}{n_{\mathrm{high}}}
\newcommand{\Horizon}{H}                    % oracle lookahead depth
\newcommand{\Sup}{S}                        % supervised steps
\newcommand{\Aopt}{\mathcal{A}^\star}
\newcommand{\Grp}{G}

\newcommand{\Cref}{C_{\mathrm{ref}}}
\newcommand{\Cbest}{C_{\mathrm{best}}}
\newcommand{\tbest}{t_{\mathrm{best}}}
\newcommand{\ratio}{\rho}
\newcommand{\eclip}{\epsilon_{\mathrm{clip}}}
\newcommand{\Credited}{\Omega}
\newcommand{\LIL}{\mathcal{L}_{\mathrm{IL}}}
\newcommand{\LRL}{\mathcal{L}_{\mathrm{RL}}}

\journal{Neural Networks}

\begin{document}

\begin{frontmatter}

%\title{Operator-Aligned Neural Improvement for Routing Problems}
\title{A First Guess Is Rarely the Final Answer: \\ Learned Local Search That Generalizes to Larger Routing Problems}

\author[1]{Andoni Irazusta Garmendia}
\ead{andoni.irazusta@ehu.eus}

\affiliation[1]{
    organization={University of the Basque Country (UPV/EHU)},
    department={Department of Computer Science and Artificial Intelligence},
    city={Donostia-San Sebastián},
    country={Spain}
}

\begin{abstract}
Neural routing solvers rarely stop at their first solution: constructed tours are usually refined by search, such as sampling or local search. Neural improvement methods learn this search as a policy that repeatedly applies local moves, but existing policies degrade sharply beyond their training size. We argue that size generalization depends on whether the information that determines the value of a move reaches the policy in a size-independent form, and identify three places where existing policies fall short: the representation of the incumbent solution, the scale of the input features, and the computation of move gains. We address all three in NICO, a learned 2-opt policy for routing that represents the incumbent by its solution edges, one token per edge without positional encodings, describes them with scale-free features, and adds the exact gain of each move to its logit, so that the network learns a correction to greedy selection. A controlled factorial study shows that edge tokens outperform node tokens with positional encodings, that the exact gains are necessary for transfer in the tested configuration, and that scale-free features help only together with them. Trained on instances with at most 100 nodes, NICO reaches a 1.1\% gap on the Euclidean TSP with 1000 nodes, against 10.7\% for greedy 2-opt and 36\% for the best prior learned policy. It also improves the tours of constructive and diffusion-based solvers, is the best improvement method on road-network costs, and on the CVRP outperforms every classical control.
\end{abstract}

% Highlights (submitted as a separate file; max. 85 characters each):
%  - Learned 2-opt policies degrade beyond their training size, behind greedy 2-opt.
%  - Size transfer needs move-value information in a size-independent form.
%  - Edge tokens expose the edges a move removes; the decoder adds its exact gain.
%  - Scale-free features help only when the decoder supplies the exact move gains.
%  - Trained on n <= 100, NICO reaches a 1.1% gap at n = 1000 (greedy 2-opt: 10.7%).


\end{frontmatter}

%===========================================================================
\section{Introduction}
\label{sec:intro}

Routing problems such as the Traveling Salesperson Problem (TSP) and the Capacitated Vehicle Routing Problem (CVRP) are core problems in logistics and a standard testbed for Neural Combinatorial Optimization (NCO)~\citep{bengio2021machine,mazyavkina2021reinforcement,cappart2023combinatorial}. Most NCO research has focused on \emph{constructive} policies, which build a solution node by node~\citep{vinyals2015pointer,kool2018attention,kwon2020pomo,luo2023neural,drakulic2023bq}. 
Yet the constructed solution is rarely the final answer: published pipelines add test-time search through sampling~\citep{kwon2020pomo}, active search~\citep{hottung2021efficient}, or 2-opt post-processing~\citep{sun2023difusco,li2024fast}, so final quality depends as much on the search as on the construction.

\emph{Neural improvement} methods make that search the object of learning~\citep{chen2019learning,lu2019learning,d2020learning,wu2021learning,ma2021learning,ma2023learning}. A policy observes the instance and the \emph{incumbent} (the current solution), selects a local move that modifies it, typically 2-opt~\citep{croes1958method}, and repeats this for a budget of steps. Since 2-opt-style neighborhoods are central to classical routing heuristics~\citep{lin1973effective,helsgaun2000effective,vidal2022hybrid}, such a policy could replace hand-crafted move selection in standalone search and in hybrid solvers.

\begin{figure}[t]
\centering
\includegraphics{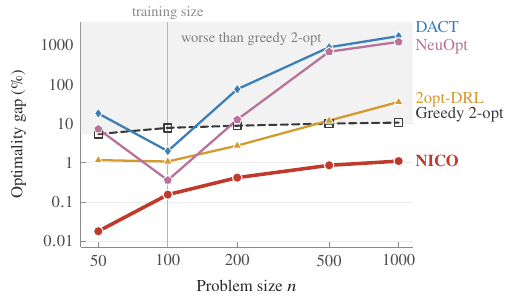}% source: images/teaser.tex, data: images/data/teaser.csv (PROVISIONAL baselines)
\caption{\textbf{Prior neural improvement methods do not generalize across problem sizes.} Optimality gap (log scale) after $10n$ steps on uniform Euclidean TSP (128 instances per size). The shaded zone marks results worse than greedy 2-opt descent at the same size; the vertical line marks the training size of the baseline checkpoints ($n=100$). All learned methods outperform greedy 2-opt at $n=100$. With these checkpoints, DACT and NeuOpt are worse than greedy 2-opt at every other size, by up to two orders of magnitude, and 2-opt DRL from $n=500$ on. NICO, the method proposed in this paper, stays well below greedy 2-opt at every size.
}
\label{fig:teaser}
\end{figure}

In practice, this promise has not been delivered. Figure~\ref{fig:teaser} evaluates the released TSP-100 checkpoints of three established methods, 2-opt DRL~\citep{d2020learning}, DACT~\citep{ma2021learning} and NeuOpt~\citep{ma2023learning}, with their official code on uniform Euclidean TSP. At their training size ($n=100$), the checkpoints behave as reported by their authors and clearly improve over greedy 2-opt descent. On larger instances they fail: at $n\ge500$, all three are worse than greedy 2-opt, although it uses no learning and a fraction of the compute.

Why do these policies fail to transfer? A 2-opt move removes two edges of the incumbent and inserts two others, and its immediate value, the change in cost, depends only on the four endpoints of the removed edges. This paper argues that the failure is not intrinsic to learned local search, and that size generalization depends on whether the information that determines the value of a move reaches the policy in a form that does not depend on the problem size. Existing policies fall short of this in three places, each of which we state as a hypothesis and test in Section~\ref{sec:ablation}:
\begin{enumerate}
    \item \textbf{Representation.} Most existing methods inherit their representation from constructive NCO, with one token per \emph{node} and the incumbent tour injected through \emph{positional encodings} (PEs)~\citep{wu2021learning,ma2021learning,ma2023learning}. A move is indexed by two node tokens, but its value also depends on their tour successors, which the encoder must retrieve by attending to ``the node at the next position''. Absolute PEs are out of distribution beyond the training size and cyclic PEs change with $n$, so this retrieval is tied to the training size. PEs also depend on the arbitrary choice of a starting node, and a single move changes the positions of $\Theta(n)$ nodes although it replaces only two edges.
    \item \textbf{Input statistics.} Absolute coordinates and raw edge lengths change their distribution with $n$: in uniform instances, the edges of a good tour have length of order $1/\sqrt{n}$. Features that a policy learns to read at $n=100$ therefore take values at $n=1000$ that it has never seen.
    \item \textbf{Implicit gains.} Even when the endpoints are available, the policy must learn to compute move gains from its embeddings and to rank $\Theta(n^2)$ candidate moves by them. As $n$ grows, the number of candidates grows quadratically while the gains of the best moves shrink, so an approximation that is accurate enough at the training size may no longer separate improving from worsening moves at larger sizes.
\end{enumerate}

\begin{figure*}[t]
    \centering
    \includegraphics[width=0.85\linewidth]{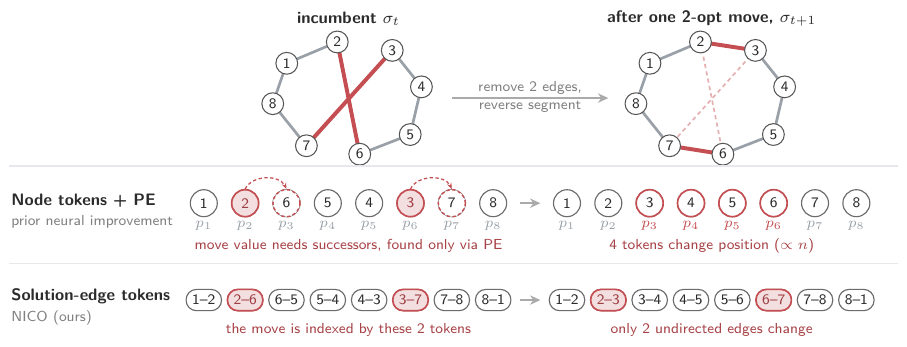}
    \caption{\textbf{The same 2-opt move seen through two state representations.} The move removes edges $\{2,6\}$ and $\{3,7\}$ from $\sol_t$ and reverses the segment between them. \emph{Top row (node tokens with PEs):} the move is indexed by the tokens of nodes 2 and 3, but its value also depends on their successors 6 and 7, which the policy can only locate through the positional encoding. After the move, every node of the reversed segment changes position, so $\Theta(n)$ tokens change. \emph{Bottom row (solution-edge tokens, NICO):} the move is indexed by the two highlighted tokens, whose endpoints determine its immediate value (Proposition~\ref{prop:edge}), and the move replaces exactly these two undirected solution edges (Section~\ref{sec:invariance}).}
    \label{fig:concept}
\end{figure*}

Figure~\ref{fig:concept} illustrates the first point on a single move. Representing edges explicitly is a natural alternative, used in constructive routing~\citep{kwon2021matrix,meng2025eformer} and heatmap methods~\citep{joshi2019efficient,hudson2022graph}, but a Transformer over \emph{all} candidate edges needs $O(n^2)$ tokens. Neural improvement does not need them: the incumbent has exactly $\Len$ edges ($\Len=n$ for the TSP, $\Len=n+\nroutes$ for a CVRP with $\nroutes$ routes), and these are precisely the objects that 2-opt removes.

We therefore propose \textbf{NICO} (\emph{Neural Improvement for Combinatorial Optimization}), which addresses each point in turn. (i)~It represents the incumbent by its $\Len$ solution edges, one token each, without PEs. The two tokens that index a 2-opt move are the two edges it removes, for every $n$; with endpoint information, the move's cost change is a function of these two tokens (Proposition~\ref{prop:edge}). (ii)~Its edge features are scale-free and invariant to translations, rotations and reflections of the instance. They omit absolute positions, and with them the information from which two tokens could compute a gain. (iii)~Its decoder supplies that information exactly: it adds the gain of each move to the move's logit, in $O(\Len^2)$ time, so the network learns a correction to greedy selection rather than the gain itself. The three choices are complementary rather than independent, and this is the main finding of our factorial study: within the tested configurations, scale-free features fail without the gain-aware decoder and give the best transfer with it. The decoder supplies the exact gain of each move, and the network, which sees only size-independent inputs, learns how to choose among the moves the gains favor. NICO is trained by imitation of an exact two-step oracle followed by critic-free group-based reinforcement learning.

\noindent Our contributions are:
\begin{enumerate}
    \item \textbf{A design analysis} of three places where the information that determines a move's value can come to depend on the problem size: the representation, the input statistics and the decoder (Section~\ref{sec:analysis}).

    \item \textbf{NICO}, an improvement policy for the TSP and the CVRP built on solution-edge tokens, scale-free features and a gain-aware decoder (Section~\ref{sec:method}), trained with oracle imitation followed by group-based RL (Section~\ref{sec:learning}).

    \item \textbf{A controlled factorial study} of the three factors over three training seeds, showing that scale-free features help only when the decoder supplies the exact gains, together with experiments that isolate what the network adds to these gains (Section~\ref{sec:ablation}).

    \item \textbf{An empirical evaluation} on Euclidean and road-network TSP and on the CVRP, against classical local search, LKH-3, Concorde and prior learned methods, including the refinement of tours from constructive and diffusion-based solvers (Sections~\ref{sec:exp_tsp}--\ref{sec:exp_cvrp}).
\end{enumerate}

The remainder of the paper is organized as follows: Section~\ref{sec:related} reviews related work, Sections~\ref{sec:prelim}--\ref{sec:learning} present the setting, analysis and method, Sections~\ref{sec:setup}--\ref{sec:exp_cvrp} report experiments, and Sections~\ref{sec:discussion}--\ref{sec:conclusion} conclude.

%===========================================================================
\section{Related Work}
\label{sec:related}
%===========================================================================

\paragraph{Constructive neural solvers}
Constructive methods generate a solution autoregressively with pointer or attention models trained by supervised learning or RL~\citep{vinyals2015pointer,bello2016neural,nazari2018reinforcement,kool2018attention,kwon2020pomo}. Later work improved generalization to larger instances through heavy decoders~\citep{luo2023neural}, bisimulation-based state reductions~\citep{drakulic2023bq}, instance-conditioned adaptation~\citep{zhou2024instance}, and self-improvement~\citep{luo2024self}. Non-autoregressive methods predict edge heatmaps that are decoded by search, including graph convolutional networks~\citep{joshi2019efficient}, DIMES~\citep{qiu2022dimes} and diffusion models~\citep{sun2023difusco,li2024fast}. Many of these pipelines use test-time search: sampling and augmentation~\citep{kwon2020pomo,kim2022symnco}, active search~\citep{hottung2021efficient}, or 2-opt post-processing~\citep{sun2023difusco,li2024fast}. Recent test-time scaling frameworks interleave neural generation with problem-specific disruption or reconstruction operators~\citep{gensco,maskco}. In GenSCO, for example, randomized 2-opt is one of the operators. These operators are applied without a learned policy, which is the component NICO learns.

\paragraph{Neural improvement methods}
Neural improvement methods learn to select local modifications of an incumbent solution. \citet{chen2019learning} learned region-picking and rewriting policies. \citet{lu2019learning} learned to select among classical operators for the CVRP. \citet{hottung2020neural} learned repair operators within large neighborhood search. \citet{d2020learning} learned 2-opt move selection with a pointer-attention policy. \citet{wu2021learning} proposed a self-attention improvement policy for the TSP and CVRP that encodes the incumbent through node positions. DACT~\citep{ma2021learning} separated node and positional embeddings and introduced a cyclic positional encoding. N2S~\citep{ma2022efficient} and NeuOpt~\citep{ma2023learning} extended this line to pickup-and-delivery problems and to flexible $k$-opt moves. The recent Transformer-based routing methods in this line use node tokens and inject the incumbent solution mainly through positional information; as discussed by \citet{ma2021learning}, some earlier policies instead add predecessor and successor information to the node features.

\paragraph{Learning-guided classical local search}
A complementary line uses neural predictions to guide classical heuristics rather than to select moves directly. NeuroLKH~\citep{xin2021neurolkh} learns candidate sets for LKH~\citep{helsgaun2000effective,helsgaun2017extension}. \citet{hudson2022graph} predict edge regret to guide local search. \citet{fu2021generalize} combine a small pretrained model with Monte Carlo tree search. These methods predict properties of candidate edges of the complete graph, whereas NICO learns a policy over the edges of the incumbent solution.

\paragraph{Edge-centric representations}
Edges are the natural decision variables of routing, and several works represent them explicitly. MatNet~\citep{kwon2021matrix} updates the representations of the two node sets of a weighted bipartite graph and incorporates the matrix entries into the attention scores. EFormer~\citep{meng2025eformer} combines matrix-aware preprocessing, a graph encoder and a node encoder for constructive vehicle routing, and RRNCO~\citep{son2025neural} learns from real-world distance matrices. These approaches inject information about the candidate edges of the \emph{instance} into node-level computation. NICO instead uses the $O(n)$ edges of the \emph{current solution} as tokens; these edges are both the carrier of the state and the arguments of the action.

\paragraph{Positional encodings and size generalization}
PEs are the standard way to convey order in Transformers~\citep{vaswani2017attention}. In graphs, they encode structural positions~\citep{dwivedi2020generalization,dwivedi2022graph}. Poor generalization across sizes is a recognized problem of neural routing solvers~\citep{joshi2020learning,zhou2023towards}. Our analysis points to one candidate mechanism for improvement policies that observe the incumbent solution only through PEs: the meaning of the PE changes with $n$.

%===========================================================================
\section{Preliminaries}
\label{sec:prelim}
%===========================================================================

\subsection{Routing problems}
\label{sec:problems}

\paragraph{TSP}
An instance $\inst$ consists of a node set $\Nodes=\{1,\dots,n\}$ with features $\nfeat{v}$ (coordinates $\nfeat{v}\in[0,1]^2$ in the Euclidean case) and symmetric costs $\cost{u}{v}=\cost{v}{u}\ge 0$ for $u,v\in\Nodes$. A solution is a permutation $\sol=(\sol_1,\dots,\sol_n)\in\Perm$, where $\Perm$ denotes the set of permutations of $\Nodes$. It is read as a cyclic sequence, with the convention $\sol_{n+1}\equiv\sol_1$, and its cost is
\begin{equation}
\Cost(\sol\mid\inst)=\sum_{k=1}^{n} \cost{\sol_k}{\sol_{k+1}}.
\label{eq:cost}
\end{equation}

\paragraph{CVRP}
A CVRP instance adds a depot, denoted node $0$, a demand $\dem{v}>0$ for each customer $v\in\Nodes$, and a vehicle capacity $\Cap$. A solution is a set of routes that start and end at the depot, visit every customer exactly once, and serve a total demand of at most $\Cap$ each. Following~\citet{wu2021learning} and~\citet{ma2021learning}, we write a solution with at most $\nroutes$ routes as a \emph{giant tour}: a cyclic sequence $\sol=(\sol_1,\dots,\sol_\Len)$ of length $\Len=n+\nroutes$ that contains every customer once and $\nroutes$ copies of the depot. Each depot copy closes one route and opens the next, and consecutive depot copies encode empty routes. The cost is again given by Eq.~\eqref{eq:cost} with $n$ replaced by $\Len$.

\paragraph{Solution edges}
In both problems, we define the (directed) solution edges of $\sol$ as
\begin{equation}
\edge{k}=(\sol_k\rightarrow\sol_{k+1}),\qquad k=1,\dots,\Len,
\label{eq:edges}
\end{equation}
and write $\Edges(\sol)=\{\{\sol_k,\sol_{k+1}\}\}_{k=1}^{\Len}$ for the multiset of the corresponding undirected edges. $\Edges(\sol)$ determines the cycle, and hence the solution, uniquely.

\subsection{The 2-opt neighborhood}
\label{sec:2opt}

A 2-opt move $\move{i}{j}$ with $1\le i<j\le \Len$, $j\ge i+2$ and $(i,j)\neq(1,\Len)$ removes the non-adjacent edges $\edge{i}$ and $\edge{j}$, reverses the segment $(\sol_{i+1},\dots,\sol_j)$, and reconnects the solution with the edges $\{\sol_i,\sol_j\}$ and $\{\sol_{i+1},\sol_{j+1}\}$~\citep{croes1958method}. For symmetric costs, the change in cost is
\begin{equation}
\dcost{i}{j}(\sol)= \cost{\sol_i}{\sol_j}+\cost{\sol_{i+1}}{\sol_{j+1}}-\cost{\sol_i}{\sol_{i+1}}-\cost{\sol_j}{\sol_{j+1}},
\label{eq:delta}
\end{equation}
because every edge inside the reversed segment keeps its cost. There are $\Len(\Len-3)/2$ admissible moves. In the CVRP, 2-opt on the giant tour produces an intra-route 2-opt move when $\edge{i}$ and $\edge{j}$ belong to the same route, and an inter-route exchange of route heads and tails (a 2-opt$^\ast$-type move with reversal) otherwise. In the latter case, feasibility requires checking the capacities of the two routes that change.

\subsection{Neural improvement as a Markov decision process}
\label{sec:mdp}

We model neural improvement as an episodic MDP with a budget of $\Budget$ steps, $t\in\{0,\dots,\Budget-1\}$. The budget may differ between training and inference. A state $s_t=(\inst,\sol_t,\hist_t)$ consists of the instance $\inst$, the incumbent solution $\sol_t$, and a short search history $\hist_t$. In NICO, $\hist_t$ is a first-in-first-out buffer of the last $\Bhist$ executed moves (Section~\ref{sec:arch}). An action $a_t=(i_t,j_t)\in\Adm(s_t)$ is an admissible 2-opt move. The transition $\sol_{t+1}=\move{i_t}{j_t}(\sol_t)$ is deterministic. A policy $\pol(a_t\mid s_t)$ is trained on instances drawn from a distribution $\Dist$ so that it transfers to unseen instances, including instances of other sizes. We are interested in the best cost found within the budget, $\min_{t\le \Budget}\Cost(\sol_t\mid\inst)$.

%===========================================================================
\section{Three Sources of Size Dependence}
\label{sec:analysis}
%===========================================================================

A neural improvement policy encodes the incumbent into $\Len$ tokens and scores each 2-opt move with a pairwise function $\scorer$ of the two tokens that index it, typically a scaled dot product $\langle W_Q h,W_K h'\rangle/\sqrt{\dkey}$. By Eq.~\eqref{eq:delta}, the immediate value of $\move{i}{j}$ depends only on the endpoints of the two removed edges $\edge{i}$ and $\edge{j}$. This section examines three places where this information can acquire a dependence on the problem size: the tokens (Section~\ref{sec:analysis_tokens}), their features (Section~\ref{sec:analysis_features}) and the decoder (Section~\ref{sec:analysis_decoder}). Each is a hypothesis about how trained policies fail under size shift, and Section~\ref{sec:ablation} tests all three.

\subsection{Tokens: retrieval versus direct access}
\label{sec:analysis_tokens}
 
A node-based policy scores a 2-opt move from the tokens of the tail nodes $\sol_i$ and $\sol_j$ of the two removed edges. By Eq.~\eqref{eq:delta}, however, the value of the move also depends on their successors $\sol_{i+1}$ and $\sol_{j+1}$, which are a property of the solution rather than of the nodes.
 
\begin{observation}[Node tokens without solution information]
\label{prop:node}
Let node tokens $\emb{v}=\Phi(\inst)_v$ be produced by any encoder that receives the instance but not the incumbent solution, and let a pairwise scorer assign the logit $\scorer(\emb{\sol_i},\emb{\sol_j})$ to the move that removes the edges leaving $\sol_i$ and $\sol_j$. Then there exist an instance, two tours $\sol,\sol'$ and two nodes such that this move is strictly improving in $\sol$, strictly worsening in $\sol'$, and receives the same logit in both.
\end{observation}
 
\begin{proof}
Place six points on a line at $p_1=0$, $p_2=1$, $p_3=10$, $p_4=11$, $p_5=-1$, $p_6=2$, with Euclidean costs, and let $\sol=(p_1,p_3,p_5,p_2,p_4,p_6)$ and $\sol'=(p_1,p_5,p_3,p_2,p_6,p_4)$. The move that removes the edges leaving $p_1$ and $p_2$ removes $\{p_1,p_3\},\{p_2,p_4\}$ in $\sol$, with $\Delta=1+1-10-10=-18$ by Eq.~\eqref{eq:delta}, and $\{p_1,p_5\},\{p_2,p_6\}$ in $\sol'$, with $\Delta'=1+3-1-1=2$. Since $\Phi$ ignores the solution, $\emb{p_1}$, $\emb{p_2}$ and hence the logit coincide. Small perturbations off the line preserve both signs.
\end{proof}
 
The observation is not an impossibility result: it only states that node-based policies must receive the successors from somewhere. Transformer policies supply them through a positional encoding (PE) of each node's tour position~\citep{wu2021learning,ma2021learning}, so the encoder must \emph{retrieve} each successor by attending to the token at the next position. We hypothesize that this retrieval is where size dependence enters, because its meaning changes with $n$ for both common PEs:
\begin{itemize}
    \item \emph{Sinusoidal PEs}~\citep{vaswani2017attention}: positions beyond the training size are never seen, and the closing step from position $n$ to position $1$ is not the same rotation as the other steps.
    \item \emph{Cyclic PEs}~\citep{ma2021learning}, with frequencies that are multiples of $2\pi/n$: the step from one position to the next is a rotation whose angle depends on $n$, so a rule that steps one position at $n_{\mathrm{train}}$ steps about $n/n_{\mathrm{train}}$ positions at size $n$, for example ten positions at $n=1000$ for a model trained at $n=100$.
\end{itemize}
Solution-edge tokens remove the need for retrieval, because the endpoints that determine a move's value are the contents of the two tokens that index it:
 
\begin{proposition}[Pairwise move evaluation]
\label{prop:edge}
Let costs be symmetric, and let $\sol$ be a solution of length $\Len$ with an admissible move $(i,j)$.
\begin{enumerate}
    \item[(a)] If the token of $\edge{k}$ contains the features of both endpoints, $\tok{k}=(\nfeat{\sol_k},\nfeat{\sol_{k+1}})$, and $\cost{u}{v}=\delta(\nfeat{u},\nfeat{v})$ for some function $\delta$, then a single function $\scorer^\ast$, independent of $\Len$ and of $\sol$, satisfies $\scorer^\ast(\tok{i},\tok{j})=-\dcost{i}{j}(\sol)$.
    \item[(b)] For an arbitrary symmetric cost matrix, the matrix of all move gains $[-\dcost{i}{j}(\sol)]_{i,j}$ is obtained from the $\Len$ solution-edge costs and two gathers of the cost matrix, in $O(\Len^2)$ time.
\end{enumerate}
\end{proposition}
 
\begin{proof}
By Eq.~\eqref{eq:delta}, $\dcost{i}{j}$ depends only on $\sol_i,\sol_{i+1},\sol_j,\sol_{j+1}$. (a) Take $\scorer^\ast\big((u,v),(u',v')\big)=\delta(u,v)+\delta(u',v')-\delta(u,u')-\delta(v,v')$. (b) The removed costs $\cost{\sol_i}{\sol_{i+1}}$ and $\cost{\sol_j}{\sol_{j+1}}$ are the edge costs of the two tokens, and the inserted costs $\cost{\sol_i}{\sol_j}$ and $\cost{\sol_{i+1}}{\sol_{j+1}}$ are the entries of the cost matrix with rows and columns permuted by $\sol$ and by its successor map.
\end{proof}
 
Part~(b) is used by the decoder (Section~\ref{sec:analysis_decoder}). Part~(a) shows that the distinction that matters is not node versus edge as such, but whether the successor information is available to the scorer directly or must be retrieved: a node token that also contains the features of its successor carries the same information as a directed edge token. Section~\ref{sec:probe} tests this distinction with a gain-ranking probe.
 
\paragraph{Invariance and locality}
\label{sec:invariance}
The edge set has two further advantages over positional encodings. \emph{Invariance:} the multiset of undirected solution edges $\Edges(\sol)$ does not depend on the starting node or direction in which the tour is written ($2\Len$ choices), nor, in the CVRP, on the order and direction of the routes in the giant tour, whereas positional encodings change under all of these choices. \emph{Locality:} a 2-opt move replaces exactly two elements of $\Edges(\sol)$, whereas with positional encodings every node of the reversed segment changes position, $\Theta(\Len)$ nodes in expectation over uniformly drawn admissible moves.

\subsection{Features: absolute versus scale-free inputs}
\label{sec:analysis_features}

Even when each token contains the right endpoints, the values of its features can depend on $n$. For $n$ points drawn uniformly from the unit square, the optimal tour length grows as $\beta\sqrt{n}$ for a constant $\beta>0$~\citep{beardwood1959shortest}, so the edges of a near-optimal tour, like nearest-neighbor distances, have length of order $1/\sqrt{n}$: at $n=1000$ they are about three times shorter than at $n=100$. As the search approaches a local optimum, a policy trained at $n=100$ therefore sees edge lengths at $n=1000$ that lie outside its training distribution.

Rescaling an instance by $\sqrt{n/n'}$ maps point density $n$ to density $n'$, so local neighborhoods at different sizes differ mainly in scale. Features that are invariant to similarity transformations (translations, rotations, reflections and uniform scaling), which we call \emph{scale-free}, remove this difference. Ratios of lengths, such as an edge length relative to its neighboring edges or to the nearest-neighbor spacing of its endpoints, and turning angles have this property; absolute coordinates, edge directions and absolute lengths do not. Scale-free features do not make the input distribution independent of $n$: the number of tokens, the fraction of nodes near the boundary and the states visited by the search still change with the size.

Scale-free features also have a price. Without absolute positions, two tokens no longer determine the distances between their endpoints, so the condition of Proposition~\ref{prop:edge}(a) fails and the gain of a move can no longer be computed from the two tokens that index it. We therefore expect scale-free features to help only when the gains are supplied by other means, which is what the decoder of Section~\ref{sec:analysis_decoder} does.

% Proposed rewrite of Section "Decoder: learned versus exact gains", Scope and Table 1 (23 Oct).

\subsection{Decoder: learned versus exact gains}
\label{sec:analysis_decoder}

A policy that receives the right information must still compute the move gains from its embeddings and rank $\Theta(\Len^2)$ moves by them. Proposition~\ref{prop:edge}(a) shows that this computation is representable, not that a trained network performs it accurately enough at sizes it has not seen. Two effects make this harder as $n$ grows. Near a local optimum, only a few of the admissible moves improve the solution, so the number of competing moves grows quadratically while the number of good ones does not. At the same time, the gains of the best moves shrink with the edge lengths (Section~\ref{sec:analysis_features}). If the logits approximate the gains with an error that does not shrink accordingly, more and more non-improving moves receive a higher logit than the best improving move, and a ranking that is accurate at the training size can fail to identify the few good moves at larger sizes.

By Proposition~\ref{prop:edge}(b), the exact gains can instead be computed outside the network and added to the logits, in $O(\Len^2)$ time, the same cost as computing the logits themselves. Dividing them by the mean edge length of the incumbent makes this term scale-free. The network then learns a correction to a Boltzmann policy over exact gains, rather than the gains themselves (Section~\ref{sec:decoder}). Because the gains do not pass through the network, this decoder can be combined with node tokens, and with features that cannot express the gains at all, such as the scale-free features above.

%===========================================================================
\section{NICO: Representation and Policy}
\label{sec:method}
%===========================================================================

\begin{figure*}[t]
    \centering
    \includegraphics[width=0.85\linewidth]{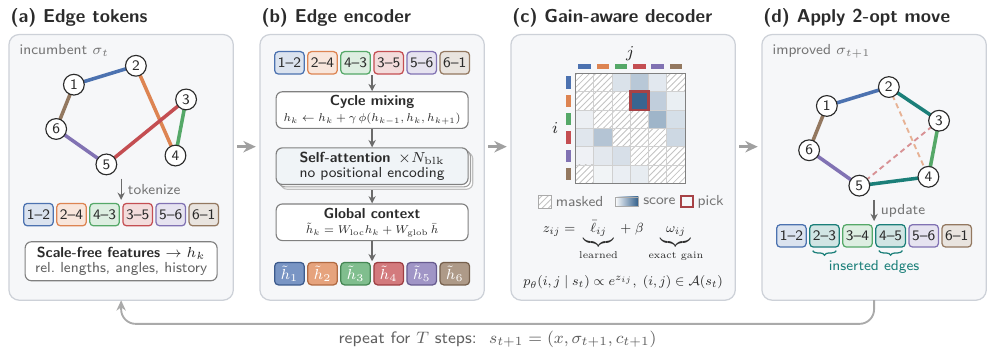}
    \caption{\textbf{One NICO improvement step.} (a)~The incumbent $\sol_t$ is represented by its $\Len$ solution edges, one token per edge, with scale-free features and no positional encoding. (b)~The encoder applies cycle mixing, $\Nblk$ self-attention blocks and a global context. (c)~The gain-aware decoder scores every 2-opt move by a learned pairwise term plus $\gbeta$ times its exact normalized gain $\gain{i}{j}$; masked entries are adjacent, tabu or (in the CVRP) capacity-infeasible pairs. (d)~The sampled move replaces exactly two undirected edges; the tokens of the reversed segment change direction, and the features of neighboring tokens and the normalization statistics are recomputed. This is repeated for $\Budget$ steps.}
    \label{fig:nico}
\end{figure*}

NICO puts the three choices of Section~\ref{sec:analysis} into one policy: solution-edge tokens, so that a move is scored from the tokens of the two edges it removes; scale-free features, so that the inputs do not shift with $n$ (Section~\ref{sec:features}); and a gain-aware decoder, so that the immediate value of each move is exact (Section~\ref{sec:decoder}). Table~\ref{tab:representations} contrasts these choices with prior node-based policies. 
What the network must learn is the part of a move's value that the immediate gain does not capture, such as how the move affects later moves. Figure~\ref{fig:nico} shows one improvement step. The encoder, decoder and learning framework are the same for every problem; only the token features, the operators and the action mask are problem-specific. We describe the model for the Euclidean TSP and then the changes for the CVRP. Exact feature definitions and hyperparameters are given in \ref{app:impl}.

\begin{table}[t]
\centering
\caption{Design choices of prior node-based improvement policies and of NICO for 2-opt on a solution with $\Len$ edges, and the section that analyzes each. NICO's tokens contain the successors (Proposition~\ref{prop:edge}(a)), and its decoder adds the exact gain to the logit (Proposition~\ref{prop:edge}(b)).}
\label{tab:representations}
\footnotesize
\setlength{\tabcolsep}{4pt}
\begin{tabularx}{\columnwidth}{@{}>{\raggedright\arraybackslash}Xcll@{}}
\toprule
 & Sec. & Node tokens + PE & NICO \\
\midrule
Successors of the cut nodes & \ref{sec:analysis_tokens} & retrieved via PEs & in the token \\
Elements changed by a move & \ref{sec:analysis_tokens} & $\Theta(\Len)$ positions & 2 edges \\
Input features & \ref{sec:analysis_features} & absolute & scale-free \\
Immediate gain $\dcost{i}{j}$ & \ref{sec:analysis_decoder} & learned & exact, in the logit \\
Tokens / attention cost & -- & $\Len$ / $O(\Len^2)$ & $\Len$ / $O(\Len^2)$ \\
\bottomrule
\end{tabularx}
\end{table}

\subsection{Edge tokens and features}
\label{sec:features}

For the solution edge $\edge{k}=(u_k\rightarrow v_k)$, with $u_k=\sol_k$ and $v_k=\sol_{k+1}$, let $\elen{k}$ be its length and let $w_k=\sol_{k-1}$ and $s_k=\sol_{k+2}$ be the neighboring nodes along the solution. Following Section~\ref{sec:analysis_features}, the tokens contain no absolute coordinates, edge directions or absolute lengths. Each token has nine scalar features in three groups:
\begin{itemize}
    \item \textbf{Relative lengths:} the length relative to the two neighboring solution edges, $\rellen{k}=\elen{k}/\tfrac12(\elen{k-1}+\elen{k+1})$; its z-score $\zs{k}$ among the edge lengths of the current solution; and its length relative to the local point spacing, $\knnlen{k}=\elen{k}/\tfrac12(\spacing{u_k}+\spacing{v_k})$, where $\spacing{v}$ is the mean distance from $v$ to its ten nearest nodes.
    \item \textbf{Turning angles:} with $\theta^{u}_k$ the turning angle at $u_k$ (from $w_k\rightarrow u_k$ to $u_k\rightarrow v_k$) and $\theta^{v}_k$ the angle at $v_k$, the features $\cos\theta^{u}_k$, $|\sin\theta^{u}_k|$, $\cos\theta^{v}_k$, $|\sin\theta^{v}_k|$ and $\sin\theta^{u}_k\sin\theta^{v}_k$. The absolute values remove the orientation, and the product retains whether both turns bend to the same side.
    \item \textbf{Search history:} the frequency $\fhist{k}$ of position $k$ among the endpoints of the last $\Bhist$ executed moves.
\end{itemize}
The features are concatenated and linearly projected to $\emb{k}\in\mathbb{R}^{\dmodel}$. The length and angle features are invariant to translations, rotations, reflections and uniform scaling of the instance, and none of the features depends on $n$ except through the solution. %Because each token distinguishes its tail $u_k$ from its head $v_k$, the representation is not invariant to reversing the direction of the tour.

\paragraph{CVRP}
For the CVRP, NICO tokenizes the edges of the giant tour (Section~\ref{sec:problems}), including depot edges, so a solution with $\nroutes$ routes has $\Len=n+\nroutes$ tokens. Two empty routes are added to the initial solution so that the policy can open a new route or empty an existing one; their depot-to-depot edges, which have zero length, are represented by a learned null token. Each token has the scale-free geometric features of the TSP and, in addition, capacity features: the normalized demands of its endpoints, depot indicators, and three loads of its route $\route(k)$, all divided by $\Cap$, namely the prefix load $\pload{k}$ served up to $u_k$, the suffix load $\rload{\route(k)}-\pload{k}$ and the residual capacity $\Cap-\rload{\route(k)}$. With these features, the capacity feasibility of a move is computable from the two tokens that index it (Section~\ref{sec:decoder}). The CVRP implementation therefore differs from the TSP one only in the capacity features, the null token and the action mask. 

\subsection{Encoder}
\label{sec:arch}

The token sequence $\{\emb{k}\}_{k=1}^{\Len}$ is cyclic, and index neighbors are solution neighbors. Instead of PEs, we inject this local structure with a residual cycle-mixing layer,
\begin{equation}
\emb{k}\leftarrow \emb{k}+\cmix\,\phi\big([\emb{k-1},\emb{k},\emb{k+1}]\big),
\label{eq:cycle}
\end{equation}
with indices modulo $\Len$, a two-layer feed-forward network $\phi$ and a learned scalar $\cmix$ initialized at $0.1$. This layer uses only relative adjacency and has no parameter that depends on $\Len$. Next, $\Nblk$ Transformer blocks without PEs produce $\{\emb{k}^{(\Nblk)}\}$, and a global summary is fused into each token,
\begin{equation}
\gsum=\frac{1}{\Len}\sum_{k=1}^{\Len}\emb{k}^{(\Nblk)},\qquad \cemb{k}=W_{\mathrm{loc}}\emb{k}^{(\Nblk)}+W_{\mathrm{glob}}\,\gsum.
\label{eq:context}
\end{equation}
Mean rather than sum pooling keeps the scale of $\gsum$ independent of $\Len$. Since self-attention without PEs is permutation-equivariant and cycle mixing is shift-equivariant, the policy is exactly equivariant to the choice of the starting position. It is not invariant to reversing the tour: the turn features and the cycle-mixing layer depend on the direction of traversal, so the invariance of the edge set to reversals (Section~\ref{sec:invariance}) does not carry over to the representation.

\subsection{Decoder and action masking}
\label{sec:decoder}

\paragraph{Gain-aware scoring}
Each move $(i,j)$ is scored from the two contextualized tokens that index it and from its exact gain,
\begin{equation}
\begin{aligned}
\logit{i}{j}&=\lmax\tanh\!\Big(\frac{\langle W_Q\cemb{i},\,W_K\cemb{j}\rangle}{\sqrt{\dkey}}\Big),\qquad \slogit{i}{j}=\tfrac12(\logit{i}{j}+\logit{j}{i}),\\
\gain{i}{j}&=\mathrm{clip}\Big(\frac{-\dcost{i}{j}(\sol)}{\Cost(\sol\mid\inst)/\Len},\,-\gmax,\,\gmax\Big),\qquad
z_{ij}=\slogit{i}{j}+\gbeta\,\gain{i}{j}.
\end{aligned}
\label{eq:logits}
\end{equation}
The learned term $\slogit{i}{j}$ is bounded by $\lmax$ and symmetrized because, for symmetric costs, $(i,j)$ and $(j,i)$ define the same move. The gain $\gain{i}{j}$ is the cost decrease in units of the mean edge length of the incumbent, which makes it scale-free (Section~\ref{sec:analysis_features}); the clip at $\gmax=10$ only affects moves that change the cost by more than ten mean edge lengths. The weight $\gbeta$ is learned and initialized at $10$, so that a move saving one tenth of a mean edge length initially receives one unit of logit more than a neutral move. The network therefore learns a correction $\slogit{i}{j}$ to a Boltzmann policy over exact gains, $\propto\exp(\gbeta\gain{i}{j})$, which serves as a learning-free baseline in Section~\ref{sec:ablation}; the variant without the gain term uses $z_{ij}=\slogit{i}{j}$. The directed convention of Eq.~\eqref{eq:edges} fixes the reconnection: $\move{i}{j}$ always inserts $\{\sol_i,\sol_j\}$ and $\{\sol_{i+1},\sol_{j+1}\}$. The policy is a masked softmax over the upper triangle,
\begin{equation}
\pol\big((i,j)\mid s_t\big)=\frac{\exp(z_{ij})\,\mathbb{I}[(i,j)\in\Adm(s_t)]}{\sum_{(i',j')\in\Adm(s_t)}\exp(z_{i'j'})}.
\label{eq:policy}
\end{equation}

\paragraph{Admissible moves}
The admissible set $\Adm(s_t)$ excludes adjacent cuts, which do not define a 2-opt move, and, as a tabu mask derived from the search history $\hist_t$, the index pairs of the last $\mtabu$ executed moves, since repeating a 2-opt move at the same positions undoes it. 

In the CVRP, $\Adm(s_t)$ also excludes capacity-infeasible moves, and their feasibility can be read from the two tokens that index the move. A move within one route keeps every route's customers and is always feasible. A move between routes $\route(i)$ and $\route(j)$ joins the parts of the two routes up to the cuts into one route and the parts after the cuts into another, leaving any complete routes in between unchanged, so it is feasible if and only if
\begin{equation}
\pload{i}+\pload{j}\le \Cap\qquad\text{and}\qquad(\rload{\route(i)}-\pload{i})+(\rload{\route(j)}-\pload{j})\le \Cap.
\label{eq:cvrp_mask}
\end{equation}
The capacity mask is therefore an $\Len\times\Len$ Boolean matrix obtained by broadcasting per-token quantities in $O(\Len^2)$ time.

\paragraph{Additional CVRP operators}
Removing two edges $\edge{i}$ and $\edge{j}$ of different routes and reconnecting the four endpoints can be done in two ways. Giant-tour 2-opt realizes one of them, which reverses the parts of the two routes between the cut edges; the other, 2-opt$^\ast$ without reversal, exchanges the route tails after $\edge{i}$ and $\edge{j}$. NICO can score both with the same tokens: each reconnection has its own bilinear head $(W_Q^{o},W_K^{o})$ and gain weight $\gbeta_o$, and a single softmax runs over the moves of both. The gain of 2-opt$^\ast$ is again a function of the four endpoints, and it is feasible if and only if $\pload{i}+\rload{\route(j)}-\pload{j}\le\Cap$ and $\pload{j}+\rload{\route(i)}-\pload{i}\le\Cap$. NICO uses both reconnections.

%===========================================================================
\section{NICO: Learning Framework}
\label{sec:learning}
%===========================================================================

\begin{figure*}[t]
    \centering
    \includegraphics[width=0.85\linewidth]{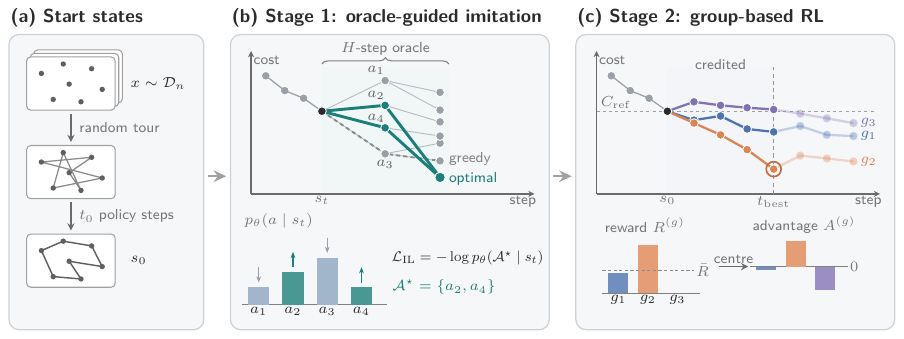}
    \caption{\textbf{Two-stage learning framework.} (a)~Both stages start from diverse states: instances of a random size, random tours, and a warmup of $t_0$ steps with a frozen snapshot of the policy. (b)~Stage~1: an exact $\Horizon$-step oracle enumerates all move sequences from $s_t$. Every first move that starts an optimal sequence belongs to the target set $\Aopt$ (here $a_2$ and $a_4$, neither of which is the greedy move $a_3$), and the policy is trained to place its mass $\pol(\Aopt\mid s_t)=\sum_{a\in\Aopt}\pol(a\mid s_t)$ on this set. (c)~Stage~2: $\Grp$ rollouts start from the same $s_0$. Each is rewarded by its improvement over $\Cref$ up to $\tbest$, the first step at which the group-best cost is reached. Rewards are centred within the group, and later steps receive no credit.}
    \label{fig:learning}
\end{figure*}

Neural improvement policies are usually trained with actor--critic RL from scratch over long horizons~\citep{wu2021learning,ma2021learning,ma2023learning}. This is sample-inefficient for two reasons. First, the action space has $O(\Len^2)$ moves, of which few improve the solution, and near a local optimum almost none do, so a randomly initialized policy rarely observes a useful reward. Second, a trajectory's return depends on hundreds of moves, many of which do not affect the final cost, so per-move credit is noisy and must be estimated by a learned critic. NICO uses two stages instead (Figure~\ref{fig:learning}): imitation of an exact short-horizon oracle, followed by critic-free group-based RL over longer horizons.

\paragraph{Training states}
Both stages draw their start states in the same way. For each mini-batch we sample a size $n\sim\mathrm{Unif}\{\nlow,\dots,\nhigh\}$, instances $\inst\sim\Dist_n$ and random initial solutions (for the CVRP, a random customer permutation split greedily into capacity-feasible routes). All instances in a mini-batch share the same $n$. We then apply $t_0\sim\mathrm{Unif}\{0,\dots,n\}$ warmup steps with a frozen snapshot $\polref$ of the policy. As the policy improves, warmup states become harder, which yields an automatic curriculum.

\subsection{Stage 1: oracle-guided imitation learning}
\label{sec:il}

Given a state $s$ and a depth $h$, the oracle enumerates all sequences of $h$ non-adjacent 2-opt moves and returns the set $\Aopt(s)$ of first moves that start a sequence with the lowest terminal cost. Several first moves can be optimal, so supervision is \emph{set-valued}. The oracle ignores the tabu mask, whereas the policy in the imitation loss is masked; states in which every oracle-optimal first move is tabu therefore have no reachable target and are skipped. The oracle costs $O(\Len^{2h})$ move evaluations per state, which is tractable at training sizes. From a warmup state $s_0$, we follow one oracle-optimal trajectory for $\Sup$ steps. At step $t$, the target set $\Aopt_t$ is computed with the remaining depth $\min(\Horizon,\Sup-t)$, and the state is advanced by a move drawn uniformly from $\Aopt_t$. We use $\Horizon=\Sup=2$ for the TSP and the CVRP. For the CVRP, the oracle enumerates only feasible moves: the capacity mask of the second move is evaluated on the solution after the first, which needs one prefix-sum pass per candidate first move and keeps the oracle exact.
The policy is trained to place its probability mass on the target sets:
\begin{equation}
\begin{aligned}
\LIL(\theta)&=-\frac{1}{\Sup}\sum_{t=0}^{\Sup-1}\log\pol(\Aopt_t\mid s_t),\\
\pol(\Aopt_t\mid s_t)&=\sum_{a\in\Aopt_t}\pol(a\mid s_t).
\end{aligned}
\label{eq:il}
\end{equation}
Unlike a cross-entropy against a single greedy move, this loss does not force the policy to break ties arbitrarily among equally good moves.

\subsection{Stage 2: group-based reinforcement learning}

For each instance $b$ in the batch, we take a warmup state $s_0^{(b)}$ and record $\Cref^{(b)}$, the best cost reached during warmup. From $s_0^{(b)}$ we run $\Grp$ independent rollouts $g=1,\dots,\Grp$ of $\Budget$ steps with the frozen policy $\polref$.
Let $\tbest^{(b)}$ be the first step at which any rollout of the group reaches the group's best cost, and set $\tbest^{(b)}=0$ if no rollout improves on $\Cref^{(b)}$. Only steps $t<\tbest^{(b)}$ receive credit.
Each rollout is scored by its best cost up to the shared horizon, relative to the warmup cost:
\begin{equation}
\begin{aligned}
\Cbest^{(b,g)}&=\min_{0\le t\le\tbest^{(b)}}\Cost\big(\sol^{(b,g)}_t\mid\inst^{(b)}\big),\\
R^{(b,g)}&=\frac{\max\big(\Cref^{(b)}-\Cbest^{(b,g)},\,0\big)}{\Cref^{(b)}}.
\end{aligned}
\label{eq:reward}
\end{equation}
The reward is centered within the group~\citep{shao2024deepseekmath} and spread evenly over the credited steps:
\begin{equation}
A^{(b,g)}_t=\frac{\mathbb{I}\big[t<\tbest^{(b)}\big]}{\tbest^{(b)}}\Big(R^{(b,g)}-\frac{1}{\Grp}\sum_{g'=1}^{\Grp}R^{(b,g')}\Big).
\label{eq:adv}
\end{equation}

Let $\Credited=\{(b,g,t):t<\tbest^{(b)}\}$ be the credited steps of the batch, $\hat A^{(b,g)}_t$ the advantages normalized to zero mean and unit variance over $\Credited$, and $\ratio^{(b,g)}_t(\theta)=\pol(a^{(b,g)}_t\mid s^{(b,g)}_t)/\polref(a^{(b,g)}_t\mid s^{(b,g)}_t)$. We minimize the PPO-style clipped loss~\citep{schulman2017proximal}
\begin{equation}
\begin{split}
\LRL(\theta)=-\frac{1}{|\Credited|}\sum_{(b,g,t)\in\Credited}\min\Big(&\ratio^{(b,g)}_t(\theta)\,\hat A^{(b,g)}_t,\\
&\mathrm{clip}\big(\ratio^{(b,g)}_t(\theta),1\pm\eclip\big)\,\hat A^{(b,g)}_t\Big).
\end{split}
\label{eq:ppo}
\end{equation}
Each collected batch is used for a single gradient step, and $\polref$ is reset to the current policy at the start of every epoch and every 20 updates. The trust region is therefore enforced by the lagging reference policy rather than by reusing data.

For the realized group, the best cost is already attained at $\tbest^{(b)}$, so later actions do not change the scores that compare the rollouts; crediting them would only add noise. A common horizon also keeps the centering exact at every credited step ($\sum_g A^{(b,g)}_t=0$), and a group in which no rollout improves contributes no gradient. The cutoff is a heuristic: it is chosen retrospectively, and later actions can influence which rollout becomes the group best.

%===========================================================================
\section{Experimental Setup}
\label{sec:setup}
%===========================================================================

The experiments address four questions, in the order of the following sections.
\textbf{(Q1)}~Which of the three factors of Section~\ref{sec:analysis} let a policy trained on $n\le100$ transfer to larger instances, what does the network add to the exact gains, and how should it be trained (Section~\ref{sec:ablation})?
\textbf{(Q2)}~How does the resulting policy compare with classical local search and prior learned methods across sizes and on TSPLIB (Section~\ref{sec:exp_tsp})?
\textbf{(Q3)}~Does NICO remain useful when the costs are road distances rather than Euclidean (Section~\ref{sec:exp_apps})?
\textbf{(Q4)}~Does the same design carry over to the capacity constraints of the CVRP (Section~\ref{sec:exp_cvrp})?

\paragraph{Data}
Training instances are generated with coordinates drawn uniformly from $[0,1]^2$ and $n\in\{20,\dots,100\}$. Model selection and the parameters of all tuned baselines use separate validation sets of 128 instances at $n=100$. The Euclidean test sets contain 128 uniform instances for each $n\in\{100,200,500,1000,2000\}$, with optimal tours from Concorde~\citep{applegate2011traveling}. At $n=2000$, Concorde did not finish within 24\,h, so the references are LKH-3 tours (default parameters). For distribution shift, we use all 43 EUC\_2D instances of TSPLIB~\citep{reinelt1991tsplib} with $100\le n\le1002$. Each is translated and scaled into the unit square, the scale of the training data, and its reference is the Concorde optimum on the unrounded Euclidean distances, so that gaps are comparable with the uniform test sets. Road-network and CVRP data are described in Sections~\ref{sec:exp_apps} and~\ref{sec:exp_cvrp}.

\paragraph{Models}
The TSP model follows Sections~\ref{sec:method}--\ref{sec:learning} with the hyperparameters of \ref{app:impl}: 100 IL epochs with $\Horizon=2$ on $n\in\{20,\dots,50\}$, then 20 RL epochs with $\Grp=20$ and $\Budget=32$ on $n\in\{20,\dots,100\}$. One training run takes about 12\,h on one RTX PRO 6000 GPU. We train \emph{three seeds} and report the mean and standard deviation over seeds; the schedule was fixed on the validation sets and the final checkpoint of each seed is evaluated.

\paragraph{Baselines}
Since NICO controls the 2-opt neighborhood, the most informative comparisons are other ways of controlling the same neighborhood. We use five groups of methods (details in \ref{app:baselines}).
\begin{itemize}
    \item \textbf{Exact and state-of-the-art heuristic solvers:} Concorde and LKH-3~\citep{helsgaun2017extension}.
    \item \textbf{Classical control of 2-opt:} greedy best-improvement 2-opt, iterated 2-opt with random 2-opt kicks, 2-opt tabu search~\citep{glover1990tabu} and simulated annealing (SA)~\citep{kirkpatrick1983optimization}, all with the same step budget as NICO; and greedy 3-opt from the 2-opt local optimum, as a descent in a larger neighborhood. Parameters are tuned on the validation sets.
    \item \textbf{Gain-Boltzmann policy:} NICO's decoder without the network, $\pi(i,j\mid s_t)\propto\exp(\gbeta\,\gain{i}{j})$ with the same tabu mask, and $\gbeta$ tuned on the $n=100$ validation set. It isolates the contribution of the learned term $\slogit{i}{j}$.
    \item \textbf{Gain-MLP:} a learned policy without an encoder that sees only per-move gain statistics and the search history. Its logit is $\gbeta\gain{i}{j}+\lmax\tanh f(x_{ij})$, where $f$ is a small MLP of the move's gain, its gap to the best gain, its gain rank, the number of improving moves and the history of its cut positions. Trained like NICO, it tests whether NICO's gain over the Gain-Boltzmann policy needs the encoder.
    \item \textbf{Learned improvement methods:} 2-opt DRL~\citep{d2020learning}, GAT-Improv~\citep{wu2021learning}, DACT~\citep{ma2021learning}, and NeuOpt~\citep{ma2023learning}, with official code and released TSP-100 checkpoints.
    \item \textbf{Constructive solvers}, used as initializers in Section~\ref{sec:refine}: nearest neighbor, POMO~\citep{kwon2020pomo} with sampling and EAS~\citep{hottung2021efficient}, LEHD~\citep{luo2023neural}, BQ-NCO~\citep{drakulic2023bq}, DIFUSCO~\citep{sun2023difusco} and Fast-T2T~\citep{li2024fast}, the last two without their built-in 2-opt.
\end{itemize}

\paragraph{Protocol}
All improvement methods start from the same initial tour of each instance, a uniformly random tour unless stated otherwise, run for $10n$ steps and return the best tour visited; greedy descents stop at their local optimum. An equal number of steps compares the quality of the move selection, not the computational budget; \ref{app:eqtime} compares the methods at equal time.
NICO samples its moves; NICO~$(\times R)$ keeps the best of $R$ independent runs. We report the mean optimality gap $100\cdot(\Cost-\Cost^\star)/\Cost^\star$. Tables of NICO over training seeds report the mean of three training seeds with one inference run each. Tables that analyze a single checkpoint report the mean of five inference runs. Time is the wall-clock time for the whole test set, with each method batched on one RTX PRO 6000 GPU or parallelized over the 12 cores of an AMD Ryzen~9 3900X CPU.

%===========================================================================
\section{What Drives Size Generalization}
\label{sec:ablation}
%===========================================================================

This section answers Q1 on the Euclidean TSP. We first test whether each representation can express move gains at all (Section~\ref{sec:probe}), then train full policies that vary the three factors of Section~\ref{sec:analysis} (Section~\ref{sec:factorial}), examine what NICO's network adds to the exact gains (Section~\ref{sec:gain_only}), and finally compare training schemes (Section~\ref{sec:learning_ablation}). The comparison with prior learned and classical methods follows in Section~\ref{sec:exp_tsp}.

\subsection{Can each representation express move gains?}
\label{sec:probe}
 
\begin{figure}[t]
    \centering
    \includegraphics{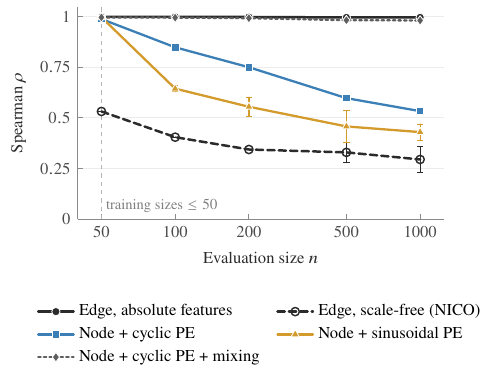}% source: images/probe.tex, data: images/data/probe.csv
    \caption{Gain-ranking probe: Spearman correlation between predicted and exact 2-opt gains over all admissible moves, for encoders with the same backbone trained to regress the gains on $n\in\{20,\dots,50\}$ and evaluated on larger tours. Mean over 3 seeds; bars show the standard deviation. Protocol in \ref{app:ablations}.}
    \label{fig:probe}
\end{figure}
 
Section~\ref{sec:analysis_tokens} argues that edge tokens give direct access to the information that determines a move's gain, whereas node tokens must retrieve it through PEs. Before training full policies, we test this directly: we train each representation to regress the exact gains of 2-opt moves on $n\in\{20,\dots,50\}$ and measure the rank correlation at larger sizes (Figure~\ref{fig:probe}; protocol in \ref{app:ablations}). With absolute features, edge tokens rank the gains almost perfectly at every size ($\rho=0.999$ at $n=50$, 0.996 at $n=1000$). Node tokens with PEs fit the training sizes but degrade as $n$ grows (cyclic PE: 0.988 to 0.533; sinusoidal PE: 0.997 to 0.430); cycle mixing, which lets each node token read its tour neighbors, recovers most of the loss (0.997 to 0.981). Edge tokens with NICO's scale-free features reach only $\rho=0.53$ at $n=50$ and 0.29 at $n=1000$: they carry no coordinates, so the cross-distances on which a gain depends cannot be computed from two tokens. The probe thus separates the two properties that NICO's tokens combine. Edge tokens make the gain of a move a function of the two tokens that index it, provided the tokens contain the endpoint positions (Proposition~\ref{prop:edge}(a)); NICO's scale-free features remove exactly this information to make the inputs size-independent. A policy with NICO's features therefore cannot obtain the gains from its tokens and must take them from the gain term of its decoder. The factorial study tests this prediction with full policies.
 
\subsection{Controlled factorial study}
\label{sec:factorial}
 
\paragraph{Design}
We train policies that differ in the factors of Section~\ref{sec:analysis} and share everything else: the Transformer blocks, global context and decoder of Sections~\ref{sec:arch}--\ref{sec:decoder}, the 2-opt action space, the tabu mask, and the training scheme and budget of the main model (Section~\ref{sec:setup}). Every configuration is trained with three seeds.
\begin{itemize}
    \item \textbf{Tokens:} node tokens with cyclic positional encodings of the tour position (node+PE), as in DACT and NeuOpt, or solution-edge tokens. With node tokens, move $(i,j)$ removes the edges that leave $\sol_i$ and $\sol_j$, so the action space is the same.
    
    \item \textbf{Features} (edge tokens only): absolute features (endpoint coordinates and raw edge lengths) or NICO's scale-free features (Section~\ref{sec:features}). Node tokens receive node coordinates, so node+PE+mix and node+mix differ from edge tokens with absolute features only in the token type and the PE.

    \item \textbf{Normalization} (edge tokens only): scale-free features change both which quantities the network sees and their scale. The \emph{raw-scale} control separates the two: it receives the same quantities unnormalized, so comparing it with the scale-free features isolates the normalization.

    \item \textbf{Decoder:} the learned bilinear term alone, or the gain-aware decoder of Eq.~\eqref{eq:logits}.
\end{itemize}
The configuration with edge tokens, scale-free features and the gain term is NICO itself. Each hypothesis corresponds to one comparison, made both without and with the gain term: \emph{representation}, node+PE against edge tokens with absolute features; \emph{input statistics}, absolute against scale-free features, with raw-scale features separating the choice of quantities from their normalization; \emph{implicit gains}, the upper against the lower block of Table~\ref{tab:factorial}.

\begin{table}[t]
\centering
\caption{Controlled factorial study: mean gap (\%) to Concorde, mean${}\pm{}$sd over three training seeds, trained on $n\le100$. Node variants use absolute features. PE: cyclic positional encoding, without cycle mixing as in DACT and NeuOpt; +mix: the cycle-mixing layer of Eq.~\eqref{eq:cycle}; Node+mix: mixing and no PE. Raw-scale: the quantities of the scale-free features without normalization. The shaded row is NICO.}
\label{tab:factorial}
\footnotesize
\setlength{\tabcolsep}{3pt}
\begin{tabular}{@{}llll@{}}
\toprule
 & $n=100$ & $n=500$ & $n=1000$ \\
\midrule
\multicolumn{4}{@{}l}{\emph{Without gain term}} \\
\quad Node+PE          & 0.428\sd{.028} & 254\sd{26}   & 747\sd{106} \\
\quad Node+PE+mix      & 0.249\sd{.037} & 20.0\sd{7.2} & 51.4\sd{18.2} \\
\quad Edge, absolute   & 0.164\sd{.028} & 6.98\sd{.84} & 22.8\sd{5.9} \\
\quad Edge, raw-scale  & 1.54\sd{.10}   & 109\sd{7}    & 282\sd{55} \\
\quad Edge, scale-free & 1.52\sd{.19}   & 514\sd{45}   & 1624\sd{699} \\
\multicolumn{4}{@{}l}{\emph{With gain term}} \\
\quad Node+PE          & 0.536\sd{.051} & 4.06\sd{.07} & 4.96\sd{.11} \\
\quad Node+PE+mix      & 0.265\sd{.013} & 3.73\sd{.17} & 4.76\sd{.13} \\
\quad Node+mix         & 0.189\sd{.060} & 3.90\sd{.30} & 5.04\sd{.22} \\
\quad Edge, absolute   & 0.127\sd{.016} & 2.02\sd{.11} & 3.10\sd{.13} \\
\quad Edge, raw-scale  & 0.146\sd{.015} & 1.48\sd{.11} & 2.55\sd{.22} \\
\rowcolor{rowshade}\quad Edge, scale-free & 0.139\sd{.032} & 0.821\sd{.041} & 1.13\sd{.07} \\
\midrule
Gain-MLP (no encoder)       & 1.85\sd{.05} & 6.01\sd{1.82} & 9.04\sd{.40} \\
Gain-Boltzmann (no network) & 2.76 & 4.57 & 5.14 \\
Greedy 2-opt                & 7.85 & 10.11 & 10.68 \\
\bottomrule
\end{tabular}
\end{table}
 
%\begin{figure}[h]
%    \centering
    %\includegraphics[width=\columnwidth]{images/factorial.pdf}% source: images/factorial.tex, data: images/data/factorial.csv
    %\caption{Size transfer of the six configurations of the factorial study (Table~\ref{tab:factorial}): mean gap (log scale) against the test size, trained on $n\le100$. Color: tokens and features; solid: with the gain term; dashed: without. The Gain-Boltzmann policy (no network) and greedy 2-opt are shown for reference.}
    %\label{fig:factorial}
%\end{figure}
 
\paragraph{Results}
Table~\ref{tab:factorial} supports all three hypotheses, but the three factors interact rather than add up.

\emph{The gain term is necessary for transfer.} Without it, every configuration fails beyond the training sizes: at $n=1000$, the gaps range from 22.8\% to 1624\%, all above greedy 2-opt (10.7\%). With it, every configuration stays below greedy 2-opt at every size and within about 5\% of the optimum at $n=1000$.
 
\emph{Edge tokens help with and without the gain term.} Without it, node+PE fails already at n = 500 (254\%), as the probe predicts; cycle mixing reduces its gap at $n=1000$ from 747\% to 51.4\%, but edge tokens reach 22.8\%. With the gain term, edge tokens reduce the gap at $n=1000$ to 3.10\%, against 4.76\% for node+PE+mix, 5.04\% for node+mix and 4.96\% for node+PE, which is no better than the Gain-Boltzmann policy (5.14\%). Access to the tour successors thus explains much of the difference without the gain term, but little of it with the gain term, where node tokens stay close to the Gain-Boltzmann policy with or without cycle mixing.

 \emph{Scale-free features help only together with the gain term, and most of their benefit comes from the normalization.} With the gain term, the scale-free features reduce the gap at $n=1000$ from 3.10\% (absolute) to 1.13\%; the raw-scale control, which contains the same quantities without normalization, reaches 2.55\%. Without the gain term, the scale-free features give the worst configuration (1624\%), because, as the probe shows, they do not contain the information from which a gain could be computed; the raw-scale control, which lacks the endpoint positions as well, also fails (282\%).

This interaction is the main finding. The decoder supplies the gains exactly, which frees the features to discard the size-dependent quantities that the gains are made of; this is the division of labor predicted in Section~\ref{sec:analysis_features}. The configuration that combines all three choices is the best within this study.
 
\subsection{What the learned term adds to exact gains}
\label{sec:gain_only}

NICO's logit for a move is the scaled gain $\gbeta\gain{i}{j}$ plus the learned term $\slogit{i}{j}$; the Gain-Boltzmann policy uses the first part alone. Comparing the two therefore isolates what the network learns. The gain coefficients are also nearly equal: NICO learns $\gbeta=5.15$, close to the value $\gbeta=5$ tuned for the Gain-Boltzmann policy on validation instances, so the two policies differ essentially only in $\slogit{i}{j}$.

\begin{table}[h]
\centering
\caption{NICO against the Gain-Boltzmann policy ($\gbeta=5$), which has the same gain term, tabu mask and budget but no network. Final gaps from Table~\ref{tab:factorial}; move statistics over the search phase ($t\ge n$), 32 test instances, $10n$ steps from random tours. Top-10: the ten admissible moves with the largest gain in the current state. Lift: mean of $\slogit{i}{j}$ on the top-10 minus its mean on the other moves, in standard deviations of $\slogit{i}{j}$.}
\label{tab:gain_only}
\footnotesize
\setlength{\tabcolsep}{3pt}
\begin{tabular}{@{}>{\raggedright\arraybackslash}p{3.7cm}rrrrrr@{}}
\toprule
& \multicolumn{3}{c}{NICO} & \multicolumn{3}{c}{Gain-Boltzmann} \\
\cmidrule(lr){2-4}\cmidrule(l){5-7}
$n$ & 100 & 500 & 1000 & 100 & 500 & 1000 \\
\midrule
Final gap (\%) & 0.14 & 0.82 & 1.13 & 2.76 & 4.57 & 5.14 \\
Improving moves (\%) & 55.2 & 56.7 & 56.3 & 51.3 & 52.1 & 52.4 \\
Best-gain move chosen (\%) & 29.9 & 26.9 & 24.9 & 12.8 & 8.0 & 6.7 \\
Chosen move in top-10 (\%) & 62.3 & 57.4 & 54.8 & 61.5 & 34.0 & 26.7 \\
sd of $\slogit{i}{j}$ / sd of $\gbeta\gain{i}{j}$ & 0.19 & 0.33 & 0.43 & 0 & 0 & 0 \\
Corr.\ of $\slogit{i}{j}$ and $\gain{i}{j}$, all moves & 0.03 & 0.01 & 0.01 & -- & -- & -- \\
Corr.\ of $\slogit{i}{j}$ and $\gain{i}{j}$, top-10 & 0.64 & 0.69 & 0.66 & -- & -- & -- \\
Lift of $\slogit{i}{j}$ on the top-10 (sd) & 0.39 & 0.72 & 0.98 & -- & -- & -- \\
\bottomrule
\end{tabular}
\end{table}

Table~\ref{tab:gain_only} shows that the two policies differ in which moves they take. NICO takes the best-gain move in 25--30\% of the steps, the Gain-Boltzmann policy in only 7--13\%, and the difference grows with $n$: at $n=1000$, 55\% of NICO's moves are among the ten with the largest gain, against 27\% for the Gain-Boltzmann policy, whose probability is spread over $\Theta(\Len^2)$ competitors. The learned term causes this concentration. On the ten moves with the largest gain, it is correlated with the gain (0.64--0.69) and higher than on the other moves by 0.4 to 1.0 standard deviations, a lift that grows with $n$. The network thus sharpens the gain ranking at its top, which is where the Boltzmann weights lose resolution as the number of moves grows.

\subsection{What each training stage contributes}
\label{sec:learning_ablation}

\begin{figure}[t]
    \centering
    \includegraphics{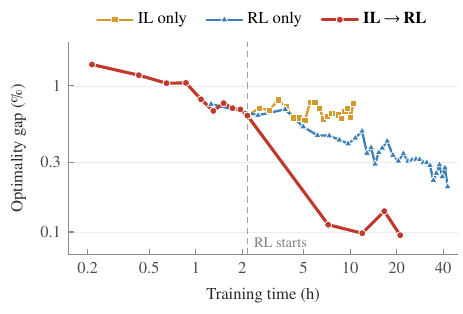}% source: images/il_rl.tex, data: images/data/il_rl_*.csv
    \caption{IL only, RL only and IL followed by RL (NICO): mean gap of each saved checkpoint at $n=100$ (128 instances, one sampled run of $10n$ steps) against training time on one RTX PRO 6000 GPU; the dashed line marks the switch from IL to RL. IL only continues the IL stage of NICO; RL only starts from a random initialization.}
    \label{fig:il_rl}
\end{figure}

NICO is trained by imitation of a two-step oracle followed by group-based RL (Section~\ref{sec:learning}). Figure~\ref{fig:il_rl} shows that both stages are needed. Imitation learns quickly but plateaus: after its 100 epochs, the policy ends at 0.75\% at $n=100$ and 2.28\% at $n=1000$, and continuing it to 500 epochs, about NICO's total training time, only reaches 0.64\% and 2.08\%. RL from a random initialization improves further but slowly, reaching 0.31\% and 1.70\% after 43 hours, three and a half times NICO's training time. Twenty RL epochs after imitation give 0.12\% and 1.14\% in 12 hours. Imitation thus provides a starting point from which a short RL stage suffices, and RL improves on what imitation alone reaches at every size. The design choices of the RL stage (late-search starts, the shared credit cutoff and the two-step oracle) are evaluated in \ref{app:training}.

%===========================================================================
\section{Results on the Euclidean TSP}
\label{sec:exp_tsp}
%===========================================================================

This section answers Q2. We first compare NICO with classical and learned improvement methods from random tours (Section~\ref{sec:perf}), and then use it to refine the tours of constructive neural solvers (Section~\ref{sec:refine}).
 
\subsection{Comparison with improvement methods}
\label{sec:perf}

\begin{table*}[t]
\centering
\caption{Euclidean TSP, trained on $n\le100$ (all other sizes and TSPLIB are zero-shot for the learned methods): mean gap (\%) to Concorde and test-set time (s). Improvement methods run $10n$ steps from the same random tours; $(\times R)$ keeps the best of $R$ runs. NICO: mean${}\pm{}$sd over training seeds. Gain-Boltzmann uses $\gbeta=5$, tuned at $n=100$. Bold: best single run among methods that control the 2-opt neighborhood. DACT and NeuOpt at $n=2000$ exceed the GPU memory.}
\label{tab:main}
\footnotesize
\setlength{\tabcolsep}{4.6pt}
 
\begin{tabular}{@{}lrrrrrrrrrrr@{}}
\toprule
& \multicolumn{2}{c}{$n=100$} & \multicolumn{2}{c}{$n=200$} & \multicolumn{2}{c}{$n=500$} & \multicolumn{2}{c}{$n=1000$} & \multicolumn{2}{c}{$n=2000$} & TSPLIB \\
\cmidrule(lr){2-3}\cmidrule(lr){4-5}\cmidrule(lr){6-7}\cmidrule(lr){8-9}\cmidrule(lr){10-11}\cmidrule(l){12-12}
Method & Gap & Time & Gap & Time & Gap & Time & Gap & Time & Gap & Time & Gap \\
\midrule
Concorde & 0.000 & 3.9 & 0.000 & 7.3 & 0.000 & 146 & 0.000 & 1440 & -- & -- & 0.000 \\
LKH-3 & 0.001 & 4.0 & 0.000 & 22.1 & 0.000 & 168 & 0.002 & 679 & 0.000 & 3110 & 0.000 \\
\midrule
\multicolumn{12}{@{}l}{\emph{Classical local search}} \\
Greedy 2-opt & 7.85 & 0.25 & 8.89 & 0.49 & 10.11 & 1.7 & 10.68 & 9.7 & 11.15 & 71 & 7.31 \\
Iterated 2-opt & 2.45 & 1.2 & 4.54 & 3.9 & 6.58 & 21.2 & 7.76 & 104 & 8.51 & 706 & 3.33 \\
Tabu search & 2.83 & 0.80 & 3.95 & 3.5 & 7.34 & 29.2 & 9.50 & 243 & 10.52 & 1910 & 3.48 \\
Simulated annealing & 2.36 & 0.97 & 3.18 & 3.8 & 4.12 & 29.4 & 4.81 & 238 & 4.94 & 1920 & 2.80 \\
Gain-Boltzmann & 2.76 & 1.2 & 3.64 & 4.6 & 4.57 & 33.2 & 5.14 & 256 & 5.45 & 2020 & 3.02 \\
Greedy 3-opt & 3.19 & 0.62 & 3.98 & 5.3 & 4.66 & 171 & 5.05 & 2784 & 5.28 & 45400 & 3.34 \\
\midrule
\multicolumn{12}{@{}l}{\emph{Learned improvement}} \\
2-opt DRL & 1.07 & 11.3 & 2.74 & 28.9 & 12.22 & 158 & 35.93 & 698 & 81.00 & 6810 & 15.90 \\
DACT & 1.89 & 16.9 & 74.94 & 70 & 897.2 & 552 & 1725.2 & 16472 & -- & -- & 100.7 \\
NeuOpt & 0.361 & 19.4 & 12.91 & 78.1 & 689.0 & 489 & 1228.5 & 19093 & -- & -- & 90.54 \\
\rowcolor{rowshade}\textbf{NICO} & \textbf{0.139\,$\pm$\,0.032} & 3.1 & \textbf{0.383\,$\pm$\,0.019} & 11.9 & \textbf{0.821\,$\pm$\,0.041} & 70.6 & \textbf{1.13\,$\pm$\,0.07} & 429 & \textbf{1.40\,$\pm$\,0.07} & 3130 & \textbf{1.07\,$\pm$\,0.33} \\
NICO $(\times 8)$ & 0.000 & 21.6 & 0.054 & 85.9 & 0.423 & 563 & 0.727 & 3551 & 1.09 & 25600 & 0.535 \\
\bottomrule
\end{tabular}
\end{table*}

Table~\ref{tab:main} shows the size shift on uniform instances. Prior learned policies do not transfer: NeuOpt and DACT are competitive at their training size but exceed 680\% at $n\ge500$, and 2-opt DRL falls behind greedy 2-opt from $n=500$. NICO stays below every classical control of 2-opt and below greedy 3-opt at every size. Simulated annealing, the strongest of our classical controls, is above it at every size.
With eight sampled runs per instance (NICO~$(\times8)$), it finds the Concorde optimum of 122 of the 128 instances at $n=100$ and of 58 at $n=200$.
Although TSPLIB instances differ from the training data in both size and spatial structure, the methods rank as on uniform instances.

\subsection{Refinement of constructive solvers}
\label{sec:refine}
 
Many neural solvers end with greedy 2-opt~\citep{sun2023difusco,li2024fast}. We test whether NICO can replace this step, starting from identical constructive tours. The initializers cover a classical heuristic (nearest neighbor), two diffusion models (DIFUSCO, Fast-T2T) and two constructive models designed for size generalization (LEHD, BQ-NCO), so that NICO is also tested on starting tours that are already good at large $n$.
 
\begin{table}[t]
\centering
\caption{Refinement of constructive tours (128 uniform instances per size): mean gap (\%) to Concorde and total pipeline time (s, construction plus refinement).}
\label{tab:refine}
\footnotesize
\setlength{\tabcolsep}{2.4pt}
 
\begin{tabular}{@{}lrrrrrr@{}}
\toprule
& \multicolumn{2}{c}{$n=100$} & \multicolumn{2}{c}{$n=500$} & \multicolumn{2}{c}{$n=1000$} \\
\cmidrule(lr){2-3}\cmidrule(lr){4-5}\cmidrule(l){6-7}
Method & Gap & Time & Gap & Time & Gap & Time \\
\midrule
Nearest neighbor & 25.24 & 0.224 & 25.69 & 1.28 & 25.44 & 3.09 \\
\quad + 2-opt & 4.68 & 0.241 & 5.45 & 1.52 & 5.54 & 4.33 \\
\quad + 3-opt & 2.67 & 0.489 & 3.42 & 85.6 & 3.64 & 1531 \\
\quad + NICO & \textbf{0.117} & 3.19 & \textbf{0.721} & 79.5 & \textbf{0.967} & 567 \\
\midrule
DIFUSCO & 1.04 & 25.3 & 6.24 & 221 & 8.52 & 1133 \\
\quad + 2-opt & 0.281 & 25.3 & 1.21 & 221 & 1.60 & 1134 \\
\quad + 3-opt & 0.125 & 25.4 & 0.828 & 249 & 1.11 & 1594 \\
\quad + NICO & \textbf{0.031} & 28.3 & \textbf{0.366} & 305 & \textbf{0.538} & 1754 \\
\midrule
Fast-T2T & 0.065 & 4.19 & 3.77 & 21.4 & 5.15 & 88.8 \\
\quad + 2-opt & 0.044 & 4.2 & 0.725 & 21.4 & 1.01 & 89.1 \\
\quad + 3-opt & 0.022 & 4.26 & 0.507 & 38 & 0.722 & 389 \\
\quad + NICO & \textbf{0.004} & 7.18 & \textbf{0.208} & 109 & \textbf{0.378} & 590 \\
\midrule
LEHD & 0.459 & 1.15 & 1.63 & 5.95 & 2.98 & 27.9 \\
\quad + 2-opt & 0.435 & 1.15 & 1.33 & 5.97 & 2.24 & 28.1 \\
\quad + 3-opt & 0.386 & 1.22 & 1.14 & 28.2 & 1.77 & 677 \\
\quad + NICO & \textbf{0.052} & 4.14 & \textbf{0.540} & 89.3 & \textbf{0.873} & 590 \\
\midrule
BQ-NCO & 0.261 & 3.05 & 1.12 & 26.9 & 2.19 & 158 \\
\quad + 2-opt & 0.252 & 3.05 & 1.04 & 26.9 & 1.88 & 158 \\
\quad + 3-opt & 0.220 & 3.11 & 0.938 & 44.6 & 1.56 & 608 \\
\quad + NICO & \textbf{0.040} & 6.04 & \textbf{0.494} & 110 & \textbf{0.824} & 720 \\
\bottomrule
\end{tabular}
\end{table}
 
NICO obtains the lowest final gap for every initializer and size (Table~\ref{tab:refine}), 1.9 to 3.8 times below greedy 3-opt at $n=1000$, and it passes both greedy methods early in its budget: from Fast-T2T tours, it reaches their converged gaps after $0.1n$ steps. The two constructive models designed for size generalization start much closer to the optimum at $n=1000$ (LEHD 2.98\%, BQ-NCO 2.19\%), and NICO still reduces their gaps to 0.87\% and 0.82\%, about half of what greedy 3-opt reaches from the same tours and below NICO's own result from random tours (1.13\%). 

NICO is also a stronger test-time search than the instance-specific adaptation of constructive models. Starting from the tours of POMO's first rollout, NICO~$(\times8)$ outperforms POMO's sampling and all three variants of Efficient Active Search (EAS) at equal wall-clock time and at every size (more details in \ref{app:pomo}): at $n=100$, it solves every instance to optimality within 0.83\,s per instance, and at $n=1000$, it reaches 0.60\% within 60\,s per instance, whereas the best EAS variant ends at 50.0\%.

%===========================================================================
\section{Non-Euclidean Costs}
\label{sec:exp_apps}
%===========================================================================

In practice, routing costs are travel distances or times on a road network rather than Euclidean distances. NICO's decoder needs only a symmetric cost matrix, since the exact gains are computed from it (Proposition~\ref{prop:edge}(b)), but its features were designed for, and trained on, Euclidean instances. This section asks whether NICO remains useful when the cost model changes (Q3). The shift studied here is in the costs rather than in the size.

We use the RRNCO instances~\citep{son2025neural}, whose costs are shortest-path distances in OpenStreetMap road networks. Since 2-opt reverses a segment, its delta is exact only for symmetric costs (Section~\ref{sec:discussion}), so we optimize the symmetrized cost $\bar{d}_{uv}=\tfrac12(\cost{u}{v}+\cost{v}{u})$, and all methods, including LEHD and RRNCO, are evaluated on it against LKH-3 references. 
Each of the three test splits contains 128 instances: unseen instances from the 80 training cities (ID), instances from 20 held-out cities (OOD-city) and clustered locations (OOD-cluster). We evaluate at $n=100$, the training size, and at $n=200$.

We compare two NICO models with the same inputs. The Euclidean model is applied \emph{zero-shot} with every edge length computed from $\bar d$, the turning angles computed from the coordinates, and the exact gains computed from $\bar d$. NICO-Road is fine-tuned from it with RL on the training cities with these same features, so the input projection is transferred unchanged.

\begin{table}[t]
\centering
\caption{Road-network TSP: mean gap (\%) to LKH-3 on the symmetrized road cost (128 instances per split) and test-set time (s) at $n=200$. Splits: in-distribution (ID), held-out cities (City), clustered locations (Clust.). Improvement methods run $10n$ steps from the same random tours. Bold: best per column.}
\label{tab:road}
\footnotesize
\setlength{\tabcolsep}{4.3pt}
\begin{tabular}{@{}lrrrrrrr@{}}
\toprule
& \multicolumn{3}{c}{$n=100$} & \multicolumn{3}{c}{$n=200$} & \\
\cmidrule(lr){2-4}\cmidrule(lr){5-7}
Method & ID & City & Clust. & ID & City & Clust. & Time \\
\midrule
LKH-3 & 0.00 & 0.00 & 0.00 & 0.00 & 0.00 & 0.00 & 100 \\
Greedy 2-opt & 3.94 & 3.40 & 3.43 & 4.74 & 3.79 & 3.61 & 1.9 \\
Greedy 3-opt & 1.70 & 1.48 & 1.48 & 2.40 & 1.69 & 1.71 & 5.6 \\
Iterated 2-opt & 0.92 & 1.05 & 0.69 & 2.10 & 1.67 & 1.64 & 3.8 \\
Tabu search & 1.56 & 1.50 & 1.97 & 2.61 & 1.98 & 2.21 & 3.5 \\
Sim.\ annealing & 1.91 & 1.75 & 2.19 & 2.61 & 2.29 & 3.22 & 3.8 \\
Gain-Boltzmann & 1.53 & 1.22 & 1.51 & 2.34 & 1.85 & 2.34 & 4.6 \\
\midrule
LEHD & 16.8 & 15.9 & 29.2 & 23.0 & 21.2 & 30.7 & 1.7 \\
RRNCO & 7.02 & 7.44 & 9.13 & 11.2 & 12.1 & 12.4 & 3.5 \\
\midrule
NICO, zero-shot & 1.19 & 1.20 & 1.55 & 2.12 & 2.19 & 4.15 & 12.2 \\
\rowcolor{rowshade}NICO-Road & \textbf{0.41} & \textbf{0.42} & \textbf{0.33} & \textbf{0.75} & \textbf{0.68} & \textbf{0.77} & 11.8 \\
\bottomrule
\end{tabular}
\end{table}

After fine-tuning on road costs, NICO-Road is the best method on every split of Table~\ref{tab:road}. Its gap changes little between the in-distribution split and the two out-of-distribution splits, and the neural constructive baselines, LEHD and RRNCO, are 7--31\% above LKH-3 on this objective. 
Without fine-tuning, the Euclidean model does not transfer to road costs as it does across sizes. Since the gain term is exact for any symmetric cost, this gap comes from the features, which were learned on Euclidean geometry; a short fine-tuning on road costs is enough to close it.

%===========================================================================
\section{Results on the CVRP}
\label{sec:exp_cvrp}
 
The CVRP tests whether the approach carries over to capacity constraints with the smallest possible change. NICO is applied to the giant tour with scale-free features plus the capacity features of \ref{app:cvrp}, the same gain-aware decoder, here over both reconnections of two removed edges (giant-tour 2-opt and 2-opt$^\ast$).
 
\paragraph{Data and baselines}
Instances follow \citet{kool2018attention}: depot and customers uniform in the unit square and demands uniform in $\{1,\dots,9\}$, with capacity $\Cap=50$ for $n=100$ and, as in~\citet{luo2023neural} and~\citet{drakulic2023bq}, $\Cap=80$, $100$ and $250$ for $n=200$, $500$ and $1000$. Test sets contain 128 instances per size, and we also use all 100 X instances of CVRPLIB~\citep{uchoa2017new} ($101\le n+1\le1001$), translated and scaled into the unit square. Gaps are computed with HGS~\citep{vidal2022hybrid} with a time limit of 20, 40, 120 and 300\,s per instance for $n=100$, $200$, $500$ and $1000$, and of $60$, $120$ and $300$\,s for X instances with $n<300$, $n<600$ and larger; these references are not always optimal, so at the largest sizes the gaps understate the distance to the optimum for every method alike. 
All improvement methods start from the same Clarke--Wright savings solutions~\citep{clarke1964scheduling}, a classical deterministic construction, and run $10n$ steps. Their gap to HGS grows with the size and the capacity, from 5.9\% at $n=100$ to 14.1\% at $n=1000$, where routes hold about 50 customers; greedy 2-opt over the giant tour, which also reorders the customers within each route, removes only 1.5 points of it there, so the gap comes mostly from the assignment of customers to routes. The classical control of the same neighborhood consists of greedy giant-tour 2-opt, iterated 2-opt with capacity-feasible random kicks, tabu search and simulated annealing, and the Gain-Boltzmann policy, all tuned on validation instances at $n=100$ (\ref{app:baselines}). Learned baselines are DACT and NeuOpt with their released CVRP checkpoints.
 
\begin{table*}[t]
\centering
\caption{CVRP, trained on $n\le100$: mean gap (\%) to HGS and test-set time (s). Improvement methods start from the same savings solutions, run $10n$ steps and return the best solution visited. n/s: not supported by the released implementation; n/r: not run. HGS: its time limit, with one instance per CPU core. NICO: mean${}\pm{}$sd over three training seeds. Bold: best per column among the improvement methods.}
\label{tab:cvrp}
\footnotesize
\setlength{\tabcolsep}{3pt}
 
\begin{tabular}{@{}lrrrrrrrrrr@{}}
\toprule
& \multicolumn{2}{c}{$n=100$} & \multicolumn{2}{c}{$n=200$} & \multicolumn{2}{c}{$n=500$} & \multicolumn{2}{c}{$n=1000$} & \multicolumn{2}{c}{X} \\
\cmidrule(lr){2-3}\cmidrule(lr){4-5}\cmidrule(lr){6-7}\cmidrule(lr){8-9}\cmidrule(l){10-11}
Method & Gap & Time & Gap & Time & Gap & Time & Gap & Time & Gap & Time \\
\midrule
HGS & 0.00 & 220 & 0.00 & 440 & 0.00 & 1320 & 0.00 & 3300 & 0.00 & 1130 \\
Savings (start) & 5.87 & 0.64 & 8.55 & 2.4 & 8.00 & 15.4 & 14.12 & 64.5 & 5.48 & 12.9 \\
\midrule
\multicolumn{11}{@{}l}{\emph{Classical control of giant-tour 2-opt}} \\
Greedy 2-opt & 5.10 & 1.2 & 7.59 & 1.6 & 7.15 & 3.8 & 12.57 & 12.7 & 4.92 & 126 \\
Iterated 2-opt & 3.51 & 9.4 & 6.14 & 33.4 & 6.51 & 157 & 11.38 & 921 & 4.06 & 387 \\
Tabu search & 3.57 & 2.3 & 6.29 & 9.1 & 6.82 & 66.8 & 12.34 & 538 & 4.35 & 296 \\
Simulated annealing & 4.40 & 2.9 & 6.82 & 10.3 & 6.69 & 74.1 & 11.52 & 599 & 4.67 & 411 \\
Gain-Boltzmann & 4.50 & 3.4 & 6.98 & 12.1 & 6.84 & 81.7 & 11.85 & 644 & 4.77 & 490 \\
\midrule
\multicolumn{11}{@{}l}{\emph{Learned improvement}} \\
DACT & 4.82 & 63.5 & 8.43 & 380 & 8.00 & 3650 & n/s & & n/r & \\
NeuOpt & 1.65 & 75.5 & 8.54 & 512 & 8.00 & 5770 & n/s & & n/r & \\
NICO, 2-opt only & 1.76 & 5.6 & 3.66 & 19.7 & 4.72 & 120 & 9.33 & 669 & \textbf{2.88} & 618 \\
\rowcolor{rowshade}\textbf{NICO} & \textbf{1.52\,$\pm$\,0.05} & 7.0 & \textbf{3.39\,$\pm$\,0.06} & 24.7 & \textbf{4.37\,$\pm$\,0.08} & 169 & \textbf{8.98\,$\pm$\,0.26} & 958 & 2.89\,$\pm$\,0.05 & 787 \\
\bottomrule
\end{tabular}
\end{table*}
 
\paragraph{Results}
NICO is the best improvement method at every size (Table~\ref{tab:cvrp}). At $n=100$, the training size of all learned methods, it ends 1.52\% above HGS (mean of three training seeds; every seed is below 1.65\%), against 1.65\% for NeuOpt, whose moves are richer than giant-tour 2-opt, 4.82\% for DACT and 3.51\% for iterated 2-opt, the best classical control at every size. Beyond the training size the two prior learned policies no longer improve the savings solutions: at $n=200$ they end at 8.43\% and 8.54\% against 8.55\% for the start, and at $n=500$ they improve none of the 128 instances. NICO ends at 3.39\%, 4.37\% and 8.98\% at $n=200$, 500 and 1000, against 6.14\%, 6.51\% and 11.38\% for iterated 2-opt, and at 2.89\% on the X instances, against 4.06\%. With the same number of steps, NICO takes about as long as iterated 2-opt (7.0 against 9.4\,s at $n=100$ and 958 against 921\,s at $n=1000$), and DACT and NeuOpt take 9 to 34 times as long. Adding 2-opt$^\ast$ to giant-tour 2-opt lowers the gap at every uniform size, by 0.24--0.35 points, and leaves it unchanged on the X instances (2.89\% against 2.88\%). All gaps are larger at $n=1000$, where the savings start is 14.1\% above HGS ($\Cap=250$, long routes) and $10n$ steps recover only part of it.
 
%\paragraph{The gain term on the CVRP}
%Table~\ref{tab:cvrp_factorial} repeats the decoder factor of Section~\ref{sec:factorial} on the CVRP: NICO with 2-opt only, trained identically with and without the gain term.
 
%\begin{table}[t]
%\centering
%\caption{The gain term on the CVRP: mean gap (\%) to HGS from savings solutions, $10n$ steps.}
%\label{tab:cvrp_factorial}
%\footnotesize
%\setlength{\tabcolsep}{4pt}
%\begin{tabular}{@{}lrrrr@{}}
%\toprule
%Model & $n=100$ & $n=200$ & $n=500$ & $n=1000$ \\
%\midrule
%NICO, 2-opt only & 1.76 & 3.66 & 4.72 & 9.33 \\
%\quad without gain term & 3.28 & 7.10 & 7.63 & 13.91 \\
%\bottomrule
%\end{tabular}
%\end{table}
%Without the gain term NICO is 1.5--4.6 points worse at every size. It improves on greedy 2-opt by only 0.5 points at $n=200$, is worse than greedy 2-opt at $n=500$ and $n=1000$, and at $n=1000$ ends barely below the savings start. As on the TSP (Table~\ref{tab:factorial}), the gain term is what carries the policy to larger instances.

%===========================================================================
\section{Discussion and Limitations}
\label{sec:discussion}
%===========================================================================

\paragraph{Why representation matters for improvement}
Constructive policies select one node per step, so node tokens are their natural unit of decision. Improvement policies edit a \emph{solution}, and 2-opt-style operators act on its edges. Our results identify the representation of the incumbent solution as one contributor to the poor size generalization of learned local search, but not as a sufficient one. Edge tokens transfer better than node tokens with positional encodings whether or not the gains are supplied, and they can represent the gains when the tokens contain endpoint positions. The best transfer, however, comes from a division of the task: the decoder supplies the exact immediate gain, and the network sees only size-independent inputs and learns a correction that sharpens the ranking of the best moves by gain (Section~\ref{sec:gain_only}). Features that make the inputs size-independent discard the quantities a gain is computed from, so they work only when the gain is supplied. We expect the same principle to apply beyond 2-opt: whatever part of a move's value can be computed exactly should be given to the policy, and the network's inputs should be chosen for invariance rather than for their ability to reconstruct that value. The training scheme does not create this advantage, but both of its stages are needed to realize it. Imitation of a two-step oracle learns quickly but plateaus. A short RL stage started from it lowers the gap at every size, whereas RL from scratch does not reach the same quality even with three and a half times the training time (Section~\ref{sec:learning_ablation}).

\paragraph{Scale}
Per-step inference is quadratic in $\Len$ for two reasons. Dense self-attention costs $O(\Len^2\dmodel)$, and the 2-opt neighborhood itself contains $O(\Len^2)$ moves. The second cost is shared by every method that ranks the full 2-opt neighborhood, including greedy 2-opt. The step budget also grows with $n$.

\paragraph{Symmetric costs and operator scope}
2-opt reverses a segment of the solution. For asymmetric costs, the change in cost then depends on the entire reversed segment, which breaks Proposition~\ref{prop:edge}. For this reason we symmetrize road costs. Orientation-preserving operators, such as relocate, or-opt and non-reversing 3-opt, have deltas that are local to the removed and inserted edges. Their moves can be scored from tuples of the directed edge tokens that NICO already uses, which makes asymmetric routing a natural next step. More generally, NICO learns to \emph{control} a given neighborhood; it does not design one.

\paragraph{Supervision horizon}
The IL oracle is optimal only over $\Horizon=2$ steps. Longer horizons would provide richer targets, but the oracle cost grows as $O(\Len^{2\Horizon})$, and optimal trajectories become non-monotone.

\paragraph{Hybridization with classical solvers}
Integrating NICO \emph{within} classical search, for example as a move selector or a diversification step, is a promising direction.

%===========================================================================
\section{Conclusion}
\label{sec:conclusion}
%===========================================================================

We have argued that a learned improvement policy generalizes across problem sizes when the information that determines the value of a move reaches it in a size-independent form, and identified three places where existing policies fall short: node tokens that retrieve tour successors through size-dependent positional encodings, input features whose scale changes with $n$, and move gains that the network must approximate among a quadratically growing number of competitors. NICO addresses them with solution-edge tokens, scale-free features and a gain-aware decoder, and a controlled factorial study shows that these choices work together: scale-free features help only because the decoder supplies the exact gains. Trained on instances with at most 100 nodes, a single NICO model reaches a 1.13\% gap at $n=1000$, against 10.7\% for greedy 2-opt and 36\% for the best prior learned policy, improves the tours of constructive and diffusion-based solvers, is the best improvement method on road networks, and, on the CVRP, outperforms every classical control of the same moves while prior learned policies no longer improve their starting solutions beyond the training size.

\section*{CRediT authorship contribution statement}
\textbf{Andoni Irazusta Garmendia:} Conceptualization, Methodology, Software, Formal analysis, Investigation, Data curation, Visualization, Writing -- original draft, Writing -- review \& editing.

\section*{Declaration of competing interest}
The authors declare that they have no known competing financial interests or personal relationships that could have appeared to influence the work reported in this paper.

\section*{Data availability}
Code, trained checkpoints and test instances will be made publicly available at \url{https://github.com/TheLeprechaun25/nico-tsp}.

\section*{Acknowledgements}
This work was supported by Basque Government through the Elkartek program under the projects KK-2024/00030, KK-2025/00098, KK-2026/00055, KK-2026/00087, and the Research Groups IT2109-26 2026-2029 program; by the Ministry of Science and Innovation MCIN/AEI/10.13039/501100011033 under the projects PID2023-149195NB-I00 and PID2023-152390NB-I00.

%===========================================================================
\appendix
%===========================================================================

\section{Implementation Details}
\label{app:impl}

\subsection{Edge-token feature definitions}
\label{app:features}

For token $\edge{k}=(u_k\to v_k)$ with predecessor $w_k=\sol_{k-1}$ and successor $s_k=\sol_{k+2}$ (indices modulo $\Len$), let $\mathbf{a}_k=\nfeat{u_k}-\nfeat{w_k}$, $\mathbf{b}_k=\nfeat{v_k}-\nfeat{u_k}$ and $\mathbf{c}_k=\nfeat{s_k}-\nfeat{v_k}$, with lengths $\elen{k-1}$, $\elen{k}$ and $\elen{k+1}$. With $\guard=10^{-6}$ and the 2D cross product $\mathbf{y}\times\mathbf{y}'=y^{(1)}y'^{(2)}-y^{(2)}y'^{(1)}$, the turn features are
\[
\cos^{(u)}_k=\frac{\langle \mathbf{a}_k,\mathbf{b}_k\rangle}{\max(\|\mathbf{a}_k\|\|\mathbf{b}_k\|,\guard)},\quad
\sin^{(u)}_k=\frac{\mathbf{a}_k\times \mathbf{b}_k}{\max(\|\mathbf{a}_k\|\|\mathbf{b}_k\|,\guard)},
\]
\[
\cos^{(v)}_k=\frac{\langle \mathbf{b}_k,\mathbf{c}_k\rangle}{\max(\|\mathbf{b}_k\|\|\mathbf{c}_k\|,\guard)},\quad
\sin^{(v)}_k=\frac{\mathbf{b}_k\times \mathbf{c}_k}{\max(\|\mathbf{b}_k\|\|\mathbf{c}_k\|,\guard)}.
\]
The normalized lengths are $\rellen{k}=\elen{k}/\max(\tfrac12(\elen{k-1}+\elen{k+1}),\guard)$ and $\zs{k}=(\elen{k}-\bar d)/\max(s_d,\guard)$, where $\bar d$ and $s_d$ are the mean and standard deviation of the edge lengths of the current solution. The history feature $\fhist{k}\in[0,1]$ counts the occurrences of position $k$ among the endpoints of the last $\Bhist$ executed moves, normalized by the number of valid entries.

For road networks, coordinates are min--max normalized per instance and axis, and the turning angles are computed from them. Both NICO models of Section~\ref{sec:exp_apps} use the features of Section~\ref{sec:features} with every edge length replaced by the symmetrized road cost $\bar d_{u_kv_k}$; because the length features are ratios, the scale of $\bar d$ does not matter. No matrix or rank features are used. NICO-Road is fine-tuned from the Euclidean checkpoint with the RL stage of the main model (group size 20, horizon 32, half of the batches with late-search starts) on instances with $n\in\{20,\dots,100\}$ from the 80 training cities, 64{,}000 instances per epoch, learning rate $10^{-4}$, for up to 30 epochs.

\subsection{Architecture and hyperparameters}

Each encoder layer applies $\emb{}\leftarrow\mathrm{Norm}(\emb{}+\mathrm{MHA}(\emb{}))$ and $\emb{}\leftarrow\mathrm{Norm}(\emb{}+\mathrm{FFN}(\emb{}))$, with a two-layer ReLU FFN of hidden width $d_{\mathrm{ff}}$, LayerNorm, no dropout and no positional encodings. The default hyperparameters are $\Nblk=3$, $\dmodel=128$, $d_{\mathrm{ff}}=512$, 8 heads, $\lmax=10$, $\Bhist=16$ and $\mtabu=8$. Training uses Muon~\citep{jordan2024muon} for the weight matrices and AdamW for all other parameters, with a shared learning rate of $3\cdot10^{-4}$ decayed by $0.99$ per epoch, weight decay $0.01$ and gradient clipping at a norm of $1.0$, with 128{,}000 instances per epoch in mini-batches of 256. IL runs for 100 epochs with $\Horizon=\Sup=2$ and warmup length $t_0\sim\mathrm{Unif}\{0,\dots,n\}$; the tabu mask is applied during IL. RL then runs for 20 epochs with $\Grp=20$, $\Budget=32$ and PPO clip $\eclip=0.2$, and half of its batches start after $t_0\sim\mathrm{Unif}\{\Len,\dots,5\Len\}$ policy steps (late-search starts); the optimizer state is reset at the transition, one gradient step is taken per collected batch, and the frozen behavior policy is copied at the start of each epoch and refreshed every 20 updates in both stages. No data augmentation is used.

\subsection{CVRP features and feasibility}
\label{app:cvrp}

Each token $\edge{k}$ of route $\route(k)$ receives the normalized demands $\dem{u_k}/\Cap$ and $\dem{v_k}/\Cap$ (with $\dem{0}=0$), the depot indicators $\mathbb{I}[u_k=0]$ and $\mathbb{I}[v_k=0]$, the prefix load $\pload{k}/\Cap$, where $\pload{k}$ is the demand served by $\route(k)$ up to and including $u_k$, the suffix load $(\rload{\route(k)}-\pload{k})/\Cap$, where $\rload{r}$ is the load of route $r$, and the residual capacity $(\Cap-\rload{\route(k)})/\Cap$. An edge leaving a depot copy belongs to the route it opens ($\pload{k}=0$); an edge entering a depot copy belongs to the route it closes ($\pload{k}=\rload{\route(k)}$). The route index $\route(k)$, the number of depot copies up to position $k$, is computed in the same $O(\Len)$ prefix-sum pass as the loads. It is used by the mask but is not an input feature; the load scalars alone do not identify route membership, so the test $\route(i)=\route(j)$ is required. For 2-opt$^\ast$, edges $\edge{i}$ and $\edge{j}$ in different routes exchange their tails, so the two new routes have loads $\pload{i}+(\rload{\route(j)}-\pload{j})$ and $\pload{j}+(\rload{\route(i)}-\pload{i})$, which gives the feasibility test of Section~\ref{sec:decoder}.

\section{Baseline Details}
\label{app:baselines}

\paragraph{Classical local search}
Greedy 2-opt applies the best improving move at each step until no improving move remains. Greedy 3-opt continues from the 2-opt local optimum and evaluates all seven non-identity reconnections of every admissible triple of edges, without pruning. The remaining methods run for $10n$ steps and return the best tour visited. Iterated 2-opt runs greedy 2-opt and, at each local optimum, returns to the best tour found and applies five random 2-opt moves with segment length at most 50. Tabu search applies the best admissible move at each step, including worsening moves, with an aspiration criterion and a tenure on removed edges of 32 steps for $n\le100$ and 64 for $n\ge200$.

Simulated annealing is run as Boltzmann sampling over the whole 2-opt neighborhood rather than with single Metropolis proposals, so that one step evaluates all moves, as a NICO step does. It samples a move from $\pi(i,j)\propto\exp\!\big(-\Delta_{ij}/(T\,C/L)\big)$, where the temperature $T$ is measured in units of the mean edge length and decreases geometrically from $T_0=1$ to $T_1=0.2$, without a tabu mask. The Gain-Boltzmann policy samples from $\pi(i,j\mid s_t)\propto\exp(\gbeta\,\gain{i}{j})$ with $\gbeta=5$ and NICO's tabu mask (tenure 8). The parameters of both are the best on the $n=100$ validation set and are used at every size; tuning them at larger sizes changes the tour length by less than 0.3\% (Table~\ref{tab:tuning}).

\begin{table}[t]
\centering
\caption{Validation tuning of the Gain-Boltzmann policy and of simulated annealing: mean tour length after $10n$ steps (128 validation instances at $n=100$ and 200, 32 at $n=500$ and 1000, 16 at $n=2000$). A coarse grid was first run at $n\le200$ ($\gbeta\in\{1,2,3,4,5,6,7,10,15,20,30\}$; $T_0\in\{0.3,0.5,1,2,3,10\}$, $T_1\in\{0.01,\dots,0.5\}$); the settings shown were then run at every size. Bold: best per size.}
\label{tab:tuning}
\footnotesize
\setlength{\tabcolsep}{3pt}
\begin{tabular}{@{}lrrrrr@{}}
\toprule
& $n=100$ & $n=200$ & $n=500$ & $n=1000$ & $n=2000$ \\
\midrule
\multicolumn{6}{@{}l}{\emph{Gain-Boltzmann}} \\
$\gbeta=3$ & 7.986 & 11.094 & 17.344 & 24.309 & 34.250 \\
$\gbeta=4$ & 7.987 & \textbf{11.060} & \textbf{17.337} & \textbf{24.229} & \textbf{34.173} \\
$\gbeta=5$ & \textbf{7.967} & 11.065 & 17.345 & 24.276 & 34.191 \\
$\gbeta=6$ & 7.982 & 11.102 & 17.355 & 24.374 & 34.294 \\
$\gbeta=7$ & 8.017 & 11.109 & 17.406 & 24.364 & 34.406 \\
\midrule
\multicolumn{6}{@{}l}{\emph{Simulated annealing}} \\
$T_0=0.5$, $T_1=0.2$ & 7.968 & 11.060 & \textbf{17.209} & \textbf{24.138} & \textbf{33.963} \\
$T_0=1$, $T_1=0.1$ & 7.970 & 11.081 & 17.284 & 24.197 & 34.067 \\
$T_0=1$, $T_1=0.2$ & \textbf{7.956} & \textbf{11.043} & 17.227 & 24.143 & 34.036 \\
$T_0=1$, $T_1=0.3$ & 7.999 & 11.077 & 17.392 & 24.271 & 34.304 \\
\bottomrule
\end{tabular}
\end{table}

\paragraph{Classical control on the CVRP}
The same methods operate on the giant tour with the capacity mask of Eq.~\eqref{eq:cvrp_mask}: greedy 2-opt, tabu search and the Gain-Boltzmann policy evaluate only feasible moves, and the kicks of iterated 2-opt are five random 2-opt moves drawn uniformly among the feasible ones with segment length at most 50, so every visited solution is feasible. Their parameters were selected on 128 validation instances at $n=100$ from savings starts and confirmed at $n=200$, where the ranking is the same: tabu tenure 64 (from $\{8,16,32,64\}$), simulated annealing from $T_0=0.3$ to $T_1=0.05$ (savings solutions are already close to the optimum, so the hot schedules tuned for random TSP tours destroy them), and $\gbeta=10$ for the Gain-Boltzmann policy (from $\{3,\dots,7,10,15\}$).

\paragraph{Neural improvement methods}
2opt-DRL~\citep{d2020learning} learns 2-opt selection with a pointer-attention policy. GAT-Improv~\citep{wu2021learning} uses a self-attention encoder with node positional information and actor--critic RL; under the protocol of Table~\ref{tab:main}, DACT~\citep{ma2021learning} embeds node features and cyclic positional features in separate streams. NeuOpt~\citep{ma2023learning} learns flexible $k$-opt moves with a recurrent dual-stream decoder and guided exploration of infeasible regions. All four use node tokens with solution-order information. We run each with its own model and rollout code and change only the initial solution and the budget. The checkpoints are the released TSP-100 models. All start from the stored node order, the same random tour as every other method, run $10n$ steps with the action selection of their test code (sampling for DACT, NeuOpt and GAT-Improv). NeuOpt uses $k\le4$ and no dynamic data augmentation stall limit, so that every method gets the same number of policy steps. 2opt-DRL rounds distances internally; its best tour is re-evaluated with exact Euclidean costs. For the CVRP, DACT and NeuOpt use their released CVRP-100 checkpoints with $0.2n$ dummy depots, start from the savings solutions and run $10n$ steps; their implementations reject instances with $1.2n\ge1000$ nodes.

\section{Additional Experiments}
\label{app:results}

This appendix gives the full results of the experiments that are summarized in the main text: the training scheme and the design choices of the RL stage, anytime behavior on the TSP, and refinement against the test-time search of a constructive model.

\subsection{Training scheme and RL design choices}
\label{app:training}

Table~\ref{tab:training} gives the numbers behind Figure~\ref{fig:il_rl} at all three test sizes (Section~\ref{sec:learning_ablation}) and evaluates three design choices of the RL stage (Section~\ref{sec:learning}): late-search starts, the shared credit cutoff against full-horizon credit, and a one-step against a two-step imitation oracle. Each variant changes one choice of NICO and keeps its schedule.

\begin{table}[t]
\centering
\caption{Training scheme and RL design choices for the full model (seed 42): mean gap (\%) on the test sets of Table~\ref{tab:main} ($10n$ steps from random tours; $n=100$: mean over 5 inference seeds, sd $\le0.08$) and training schedule in epochs; time on one RTX PRO 6000 GPU where the run was made there. Variants below the line change one choice of NICO and keep its schedule. Training-seed sd of NICO: $\le0.075$ points.}
\label{tab:training}
\footnotesize
\setlength{\tabcolsep}{3pt}

\begin{tabular}{@{}lrrrlr@{}}
\toprule
Variant & $n=100$ & $n=500$ & $n=1000$ & Epochs & Hours \\
\midrule
IL + RL (NICO) & \textbf{0.121} & \textbf{0.857} & \textbf{1.144} & IL 100 + RL 20 & 12.0 \\
IL only & 0.748 & 1.848 & 2.284 & IL 100 & 2.2 \\
IL only, longer & 0.638 & 1.686 & 2.080 & IL 500 & 10.5 \\
RL only & 0.312 & 1.243 & 1.696 & RL 64 & 42.7 \\
\midrule
No late starts & 0.200 & 1.077 & 1.347 & IL 100 + RL 20 & \\
Full-horizon credit & 1.069 & 1.836 & 1.897 & IL 100 + RL 20 & \\
IL oracle $\Horizon=1$ & 0.562 & 1.975 & 2.570 & IL 100 + RL 20 & \\
\bottomrule
\end{tabular}
\end{table}

Late-search starts help at every size, and more beyond the training sizes. Crediting each move with the improvement over the full horizon instead of the shared cutoff degrades every size. The second step of the oracle matters, and more so at large $n$.

\subsection{Anytime behavior}
\label{app:anytime}

\begin{figure*}[t]
    \centering
    \includegraphics{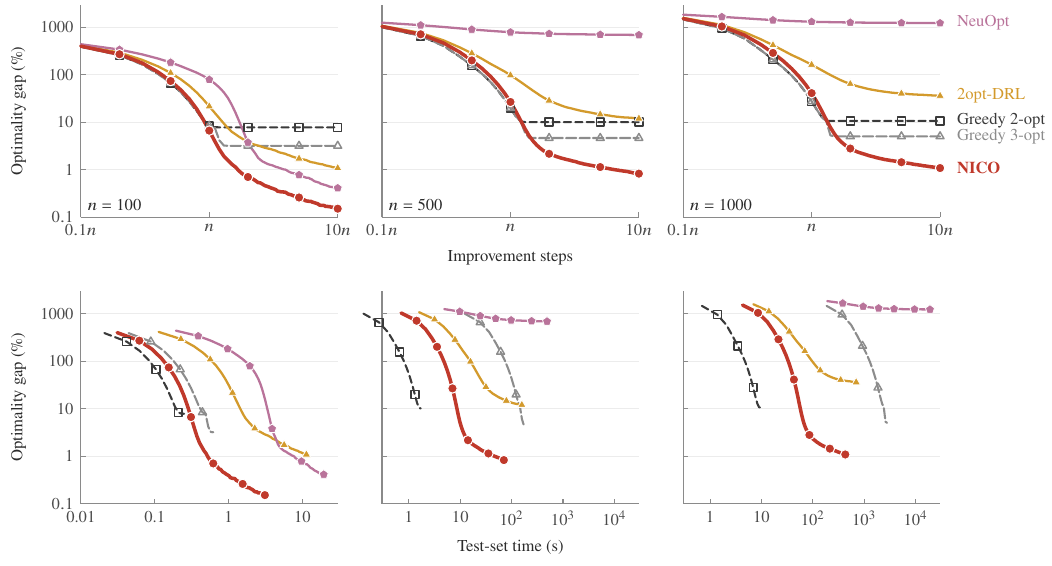}% source: images/anytime.tex, data: images/data/anytime_n*.csv
    \caption{Anytime performance from the shared random tours at $n=100$, 500 and 1000: mean optimality gap (log scale) against the number of steps (top) and against test-set time (bottom). Greedy descents stop at their local optimum. The time axis maps steps linearly onto each method's measured test-set time (Table~\ref{tab:main}); for the greedy descents, onto the steps until all instances have converged.}
    \label{fig:anytime}
\end{figure*}

Figure~\ref{fig:anytime} shows how the gap evolves within the budget. Against steps, it shows that NICO is not a greedy rule: at $n\ge500$, greedy 2-opt descends faster during the first $n$ steps, and NICO overtakes it after about $1.3n$--$1.4n$ steps and greedy 3-opt after $1.5n$--$1.6n$ steps, then keeps improving instead of stopping at a local optimum. Against time, it shows what this costs, since one NICO step is a forward pass while one greedy step is a single evaluation of the 2-opt deltas. Greedy 2-opt reaches its local optimum in 0.25, 1.7 and 9.7\,s for the test sets at $n=100$, 500 and 1000; NICO passes that gap after 0.31, 9.2 and 60\,s. Greedy 3-opt, whose steps scan a neighborhood of $\Theta(n^3)$ moves, needs 0.62, 171 and 2784\,s to reach its local optimum, while NICO passes its final gap after 0.37, 10.6 and 69\,s, 40 times sooner at $n=1000$.

\subsection{Equal-time comparisons}
\label{app:eqtime}

Table~\ref{tab:main} gives every method $10n$ steps. Table~\ref{tab:eqtime} instead gives each method NICO's measured test-set time. NICO remains below all of the baselines at every size.

\begin{table}[h]
\centering
\caption{Equal time: mean gap (\%) to Concorde (LKH-3 at $n=2000$) when every method gets NICO's test-set time of Table~\ref{tab:main}. Iterated 2-opt, tabu search, SA and Gain-Boltzmann run for the number of steps that fits in that time, from the same random tours.}
\label{tab:eqtime}
\footnotesize
\setlength{\tabcolsep}{3pt}
\begin{tabular}{@{}lrrrrr@{}}
\toprule
Method & $n=100$ & $n=200$ & $n=500$ & $n=1000$ & $n=2000$ \\
\midrule
NICO time (s)      & 3.1 & 11.9 & 70.6 & 429 & 3130 \\
\midrule
Iterated 2-opt     & 1.30 & 2.75 & 4.50 & 5.38 & 6.18 \\
Tabu search        & 2.00 & 3.11 & 7.27 & 9.50 & 10.52 \\
SA  & 1.06 & 1.94 & 2.85 & 3.59 & 3.98 \\
Gain-Boltzmann     & 1.97 & 2.80 & 3.89 & 4.59 & 4.99 \\
NICO               & 0.139 & 0.383 & 0.821 & 1.13 & 1.40 \\
\bottomrule
\end{tabular}
\end{table}

\paragraph{Refinement at equal time}
Table~\ref{tab:refine_eqtime} repeats the refinement pipelines of Table~\ref{tab:refine} with the classical pipeline given the same total time as the NICO pipeline: greedy 3-opt from the constructive tour, then iterated 2-opt for the remaining time.  NICO's gap is lower for every initializer and size. From nearest-neighbor and LEHD tours at $n=1000$, greedy 3-opt alone does not finish within NICO's time.

\begin{table}[t]
\centering
\caption{Refinement at equal time: mean gap (\%) to Concorde. Cls.: the classical pipeline (greedy 3-opt, then iterated 2-opt) given the total time of the NICO pipeline of Table~\ref{tab:refine}; its number of iterated 2-opt steps is the remaining time divided by the time per step in Table~\ref{tab:main}. --: greedy 3-opt alone takes longer than the NICO pipeline.}
\label{tab:refine_eqtime}
\footnotesize
\setlength{\tabcolsep}{3.5pt}
\begin{tabular}{@{}lrrrrrr@{}}
\toprule
& \multicolumn{2}{c}{$n=100$} & \multicolumn{2}{c}{$n=500$} & \multicolumn{2}{c}{$n=1000$} \\
\cmidrule(lr){2-3}\cmidrule(lr){4-5}\cmidrule(l){6-7}
Initializer & Cls. & NICO & Cls. & NICO & Cls. & NICO \\
\midrule
Nearest neighbor & 1.25  & \textbf{0.117} & --    & \textbf{0.721} & --    & \textbf{0.967} \\
DIFUSCO          & 0.088 & \textbf{0.031} & 0.757 & \textbf{0.366} & 1.06  & \textbf{0.538} \\
Fast-T2T         & 0.019 & \textbf{0.004} & 0.428 & \textbf{0.208} & 0.689 & \textbf{0.378} \\
LEHD             & 0.347 & \textbf{0.052} & 1.09  & \textbf{0.540} & --    & \textbf{0.873} \\
BQ-NCO           & 0.184 & \textbf{0.040} & 0.909 & \textbf{0.494} & 1.55  & \textbf{0.824} \\
\bottomrule
\end{tabular}
\end{table}

\subsection{Refinement against constructive test-time search}
\label{app:pomo}

\begin{figure*}[t]
    \centering
    \includegraphics{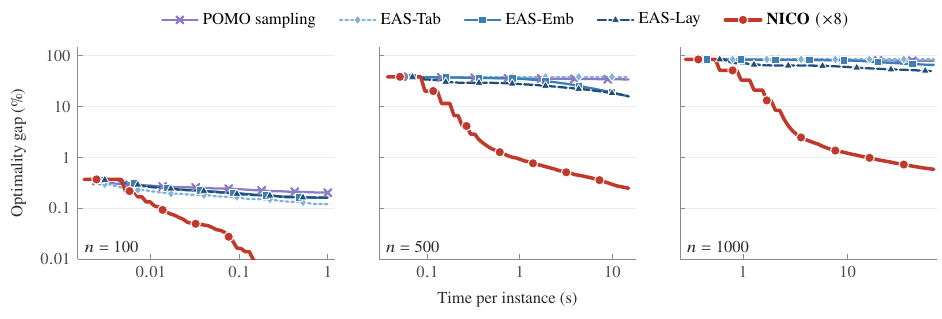}% source: images/pomo.tex, data: images/data/fig_pomo_n*.csv
    \caption{NICO as refinement of POMO tours against POMO's own test-time search: mean best-so-far gap (log scale) against wall-clock time per instance at $n=100$, 500 and 1000. POMO uses its released TSP-100 model with sampling or with the three EAS variants. NICO~$(\times8)$ refines the 8 best distinct tours of POMO's first rollout; its curve starts after that rollout. Gaps below 0.01\% are not shown.}
    \label{fig:pomo}
\end{figure*}

Figure~\ref{fig:pomo} compares NICO's refinement of POMO tours (Section~\ref{sec:refine}) with POMO's own test-time search at equal wall-clock time. At $n=100$, POMO's training size, sampling and EAS end between 0.12\% and 0.20\%, while NICO~$(\times8)$ reaches the optimum of every instance within 0.83\,s per instance. At $n=500$ and $n=1000$, POMO's first rollout is 38\% and 84\% above the optimum, and neither sampling nor instance-specific adaptation recovers (best EAS variants: 15.7\% and 50.0\%); from the same tours, NICO~$(\times8)$ reaches 0.25\% within 15\,s per instance at $n=500$ and 0.60\% within 60\,s at $n=1000$. A size-generalizing improvement policy can thus extend a constructive model that does not generalize.

\section{Ablations and Sensitivity}
\label{app:ablations}

\subsection{Search history and tabu mask}

\begin{table}[t]
\centering
\caption{Tabu tenure at inference: mean gap (\%) of one NICO checkpoint, $10n$ sampled steps from random tours ($n=100$: mean of five inference runs). $\mtabu=0$ removes the tabu mask; the history feature is kept.}
\label{tab:history}
\footnotesize
\begin{tabular}{@{}lrrr@{}}
\toprule
Tenure & $n=100$ & $n=500$ & $n=1000$ \\
\midrule
$\mtabu=0$ (no tabu mask) & 0.72 & 1.95 & 2.56 \\
$\mtabu=4$ & 0.17 & 0.93 & 1.19 \\
\rowcolor{rowshade}$\mtabu=8$ (NICO) & 0.12 & 0.86 & 1.14 \\
$\mtabu=16$ & 0.13 & 0.81 & 1.09 \\
\bottomrule
\end{tabular}
\end{table}

Table~\ref{tab:history} varies the tabu tenure at inference for a model trained with $\mtabu=8$. Without the mask the gap grows two- to sixfold, since the policy then undoes its own moves. A tenure of 16 is similar to the default; we keep $\mtabu=8$, the value used in training.

\subsection{Gain-ranking probe protocol}
The probe of Section~\ref{sec:probe} asks whether a representation can express the exact 2-opt gains and whether this transfers to larger tours, independently of search and of RL.

\paragraph{Model} Each probe uses the NICO encoder of Section~\ref{sec:arch} (embedding dimension 128, 3 layers, no search-history features) with the token type and input features under test, followed by a symmetric bilinear head $\hat{g}_{ij}=\tfrac12(s_{ij}+s_{ji})$ with $s_{ij}=(W_q\mathbf{h}_i)^\top(W_k\mathbf{h}_j)/\sqrt{128}$ over the token embeddings $\mathbf{h}$. The five representations are edge tokens with absolute features (endpoint coordinates, raw lengths); edge tokens with NICO's scale-free features (Section~\ref{sec:features}); node tokens with node coordinates and cyclic or sinusoidal PEs of the tour position; and node tokens with cyclic PEs and cycle mixing. With node tokens, move $(i,j)$ is read from the tokens of $\sol_i$ and $\sol_j$, as in the factorial study.

\paragraph{Targets and training} The target of move $(i,j)$ is its normalized gain $-\Delta_{ij}/(\Cost/n)$, the gain in units of the mean edge length of the current tour, so that targets have the same scale at every $n$. The loss is the mean squared error over all admissible moves. A training batch contains 128 instances of one size drawn uniformly from $\{20,\dots,50\}$; the tours are random tours advanced by a random number of greedy 2-opt steps, between 0 and $n$, so that states range from random to nearly 2-opt-optimal. Each probe is trained for 20{,}000 steps with AdamW (learning rate $3\cdot10^{-4}$, cosine decay, gradient clipping at 1).

\paragraph{Evaluation} For each $n\in\{50,100,200,500,1000\}$, 128 states are generated in the same way with a fixed seed, identical for all probes. We report the Spearman correlation between predicted and exact gains over the admissible moves of each state ($n(n-3)/2$ moves), averaged over states; the mean squared error and the fraction of states in which the top-ranked move is a best move are recorded as well. Every representation is trained with three seeds (0, 1, 2); Figure~\ref{fig:probe} shows the mean and standard deviation.

\paragraph{Results} Spearman $\rho$ at $n=50$ / 1000 (mean over seeds): edge tokens with absolute features 0.999 / 0.996; node + cyclic PE with mixing 0.997 / 0.981; node + cyclic PE 0.988 / 0.533; node + sinusoidal PE 0.997 / 0.430; edge tokens with scale-free features 0.532 / 0.295. The scale-free features contain no coordinates, so two edge tokens do not determine the distances between their endpoints, $\cost{\sol_i}{\sol_j}$ and $\cost{\sol_{i+1}}{\sol_{j+1}}$, on which the gain depends; the probe with these features cannot fit the gains even at the training sizes.

\bibliographystyle{elsarticle-harv}
\bibliography{references}

@inproceedings{nazari2018reinforcement,
  title={Reinforcement Learning for Solving the Vehicle Routing Problem},
  author={Nazari, Mohammadreza and Oroojlooy, Afshin and Snyder, Lawrence and Tak{\'a}{\v{c}}, Martin},
  booktitle={Advances in Neural Information Processing Systems},
  volume={31},
  year={2018}
}

@article{schulman2017proximal,
  title={Proximal Policy Optimization Algorithms},
  author={Schulman, John and Wolski, Filip and Dhariwal, Prafulla and Radford, Alec and Klimov, Oleg},
  journal={arXiv preprint arXiv:1707.06347},
  year={2017}
}

@book{applegate2011traveling,
  title={The Traveling Salesman Problem: A Computational Study},
  author={Applegate, David L. and Bixby, Robert E. and Chv{\'a}tal, Va{\v{s}}ek and Cook, William J.},
  series={Princeton Series in Applied Mathematics},
  year={2006},
  publisher={Princeton University Press}
}

@article{shao2024deepseekmath,
  title={{DeepSeekMath}: Pushing the Limits of Mathematical Reasoning in Open Language Models},
  author={Shao, Zhihong and Wang, Peiyi and Zhu, Qihao and Xu, Runxin and Song, Junxiao and Bi, Xiao and Zhang, Haowei and Zhang, Mingchuan and Li, Y. K. and Wu, Y. and Guo, Daya},
  journal={arXiv preprint arXiv:2402.03300},
  year={2024}
}

@inproceedings{d2020learning,
  title={Learning 2-opt Heuristics for the Traveling Salesman Problem via Deep Reinforcement Learning},
  author={{de O. da Costa}, Paulo R. and Rhuggenaath, Jason and Zhang, Yingqian and Akcay, Alp},
  booktitle={Proceedings of the 12th Asian Conference on Machine Learning},
  series={Proceedings of Machine Learning Research},
  volume={129},
  pages={465--480},
  year={2020},
  publisher={PMLR}
}

@inproceedings{vinyals2015pointer,
  title={Pointer Networks},
  author={Vinyals, Oriol and Fortunato, Meire and Jaitly, Navdeep},
  booktitle={Advances in Neural Information Processing Systems},
  volume={28},
  year={2015}
}

@inproceedings{bello2016neural,
  title={Neural Combinatorial Optimization with Reinforcement Learning},
  author={Bello, Irwan and Pham, Hieu and Le, Quoc V. and Norouzi, Mohammad and Bengio, Samy},
  booktitle={International Conference on Learning Representations (Workshop Track)},
  year={2017},
  note={arXiv:1611.09940}
}

@inproceedings{kwon2020pomo,
  title={{POMO}: Policy Optimization with Multiple Optima for Reinforcement Learning},
  author={Kwon, Yeong-Dae and Choo, Jinho and Kim, Byoungjip and Yoon, Iljoo and Gwon, Youngjune and Min, Seungjai},
  booktitle={Advances in Neural Information Processing Systems},
  volume={33},
  pages={21188--21198},
  year={2020}
}

@article{kirkpatrick1983optimization,
  title={Optimization by Simulated Annealing},
  author={Kirkpatrick, Scott and Gelatt, Jr., C. Daniel and Vecchi, Mario P.},
  journal={Science},
  volume={220},
  number={4598},
  pages={671--680},
  year={1983}
}

@article{helsgaun2000effective,
  title={An Effective Implementation of the {Lin--Kernighan} Traveling Salesman Heuristic},
  author={Helsgaun, Keld},
  journal={European Journal of Operational Research},
  volume={126},
  number={1},
  pages={106--130},
  year={2000}
}

@inproceedings{ma2023learning,
  title={Learning to Search Feasible and Infeasible Regions of Routing Problems with Flexible Neural k-Opt},
  author={Ma, Yining and Cao, Zhiguang and Chee, Yeow Meng},
  booktitle={Advances in Neural Information Processing Systems},
  volume={36},
  pages={49555--49578},
  year={2023}
}

@inproceedings{ma2021learning,
  title={Learning to Iteratively Solve Routing Problems with Dual-Aspect Collaborative Transformer},
  author={Ma, Yining and Li, Jingwen and Cao, Zhiguang and Song, Wen and Zhang, Le and Chen, Zhenghua and Tang, Jing},
  booktitle={Advances in Neural Information Processing Systems},
  volume={34},
  pages={11096--11107},
  year={2021}
}

@inproceedings{luo2023neural,
  title={Neural Combinatorial Optimization with Heavy Decoder: Toward Large Scale Generalization},
  author={Luo, Fu and Lin, Xi and Liu, Fei and Zhang, Qingfu and Wang, Zhenkun},
  booktitle={Advances in Neural Information Processing Systems},
  volume={36},
  pages={8845--8864},
  year={2023}
}

@inproceedings{hottung2021efficient,
  title={Efficient Active Search for Combinatorial Optimization Problems},
  author={Hottung, Andr{\'e} and Kwon, Yeong-Dae and Tierney, Kevin},
  booktitle={International Conference on Learning Representations},
  year={2022}
}

@inproceedings{chen2019learning,
  title={Learning to Perform Local Rewriting for Combinatorial Optimization},
  author={Chen, Xinyun and Tian, Yuandong},
  booktitle={Advances in Neural Information Processing Systems},
  volume={32},
  year={2019}
}

@article{cappart2023combinatorial,
  title={Combinatorial Optimization and Reasoning with Graph Neural Networks},
  author={Cappart, Quentin and Ch{\'e}telat, Didier and Khalil, Elias B. and Lodi, Andrea and Morris, Christopher and Veli{\v{c}}kovi{\'c}, Petar},
  journal={Journal of Machine Learning Research},
  volume={24},
  number={130},
  pages={1--61},
  year={2023}
}

@article{glover1990tabu,
  title={Tabu Search: A Tutorial},
  author={Glover, Fred},
  journal={Interfaces},
  volume={20},
  number={4},
  pages={74--94},
  year={1990}
}

@inproceedings{kool2018attention,
  title={Attention, Learn to Solve Routing Problems!},
  author={Kool, Wouter and van Hoof, Herke and Welling, Max},
  booktitle={International Conference on Learning Representations},
  year={2019}
}

@inproceedings{meng2025eformer,
  title={{EFormer}: An Effective Edge-based Transformer for Vehicle Routing Problems},
  author={Meng, Dian and Cao, Zhiguang and Wu, Yaoxin and Hou, Yaqing and Ge, Hongwei and Zhang, Qiang},
  booktitle={Proceedings of the Thirty-Fourth International Joint Conference on Artificial Intelligence},
  pages={8582--8590},
  year={2025},
  doi={10.24963/ijcai.2025/954}
}

@inproceedings{vaswani2017attention,
  title={Attention Is All You Need},
  author={Vaswani, Ashish and Shazeer, Noam and Parmar, Niki and Uszkoreit, Jakob and Jones, Llion and Gomez, Aidan N. and Kaiser, {\L}ukasz and Polosukhin, Illia},
  booktitle={Advances in Neural Information Processing Systems},
  volume={30},
  year={2017}
}

@article{mazyavkina2021reinforcement,
  title={Reinforcement Learning for Combinatorial Optimization: A Survey},
  author={Mazyavkina, Nina and Sviridov, Sergey and Ivanov, Sergei and Burnaev, Evgeny},
  journal={Computers \& Operations Research},
  volume={134},
  pages={105400},
  year={2021}
}

@article{joshi2019efficient,
  title={An Efficient Graph Convolutional Network Technique for the Travelling Salesman Problem},
  author={Joshi, Chaitanya K. and Laurent, Thomas and Bresson, Xavier},
  journal={arXiv preprint arXiv:1906.01227},
  year={2019}
}

@article{joshi2020learning,
  title={Learning the Travelling Salesperson Problem Requires Rethinking Generalization},
  author={Joshi, Chaitanya K. and Cappart, Quentin and Rousseau, Louis-Martin and Laurent, Thomas},
  journal={Constraints},
  volume={27},
  number={1--2},
  pages={70--98},
  year={2022}
}

@inproceedings{dwivedi2020generalization,
  title={A Generalization of Transformer Networks to Graphs},
  author={Dwivedi, Vijay Prakash and Bresson, Xavier},
  booktitle={AAAI Workshop on Deep Learning on Graphs: Methods and Applications},
  year={2021},
  note={arXiv:2012.09699}
}

@article{bengio2021machine,
  title={Machine Learning for Combinatorial Optimization: A Methodological Tour d'Horizon},
  author={Bengio, Yoshua and Lodi, Andrea and Prouvost, Antoine},
  journal={European Journal of Operational Research},
  volume={290},
  number={2},
  pages={405--421},
  year={2021}
}

@article{wu2021learning,
  title={Learning Improvement Heuristics for Solving Routing Problems},
  author={Wu, Yaoxin and Song, Wen and Cao, Zhiguang and Zhang, Jie and Lim, Andrew},
  journal={IEEE Transactions on Neural Networks and Learning Systems},
  volume={33},
  number={9},
  pages={5057--5069},
  year={2022}
}

@inproceedings{lu2019learning,
  title={A Learning-based Iterative Method for Solving Vehicle Routing Problems},
  author={Lu, Hao and Zhang, Xingwen and Yang, Shuang},
  booktitle={International Conference on Learning Representations},
  year={2020}
}

@inproceedings{hottung2020neural,
  title={Neural Large Neighborhood Search for the Capacitated Vehicle Routing Problem},
  author={Hottung, Andr{\'e} and Tierney, Kevin},
  booktitle={ECAI 2020 -- 24th European Conference on Artificial Intelligence},
  series={Frontiers in Artificial Intelligence and Applications},
  volume={325},
  pages={443--450},
  year={2020},
  publisher={IOS Press}
}

@inproceedings{sun2023difusco,
  title={{DIFUSCO}: Graph-based Diffusion Solvers for Combinatorial Optimization},
  author={Sun, Zhiqing and Yang, Yiming},
  booktitle={Advances in Neural Information Processing Systems},
  volume={36},
  pages={3706--3731},
  year={2023}
}

@inproceedings{li2024fast,
  title={{Fast T2T}: Optimization Consistency Speeds Up Diffusion-Based Training-to-Testing Solving for Combinatorial Optimization},
  author={Li, Yang and Guo, Jinpei and Wang, Runzhong and Zha, Hongyuan and Yan, Junchi},
  booktitle={Advances in Neural Information Processing Systems},
  volume={37},
  pages={30179--30206},
  year={2024}
}

@inproceedings{son2025neural,
  title={{RRNCO}: Towards Real-World Routing with Neural Combinatorial Optimization},
  author={Son, Jiwoo and Zhao, Zhikai and Berto, Federico and Hua, Chuanbo and Cao, Zhiguang and Kwon, Changhyun and Park, Jinkyoo},
  booktitle={International Conference on Learning Representations},
  year={2026}
}

@inproceedings{fu2021generalize,
  title={Generalize a Small Pre-trained Model to Arbitrarily Large {TSP} Instances},
  author={Fu, Zhang-Hua and Qiu, Kai-Bin and Zha, Hongyuan},
  booktitle={Proceedings of the AAAI Conference on Artificial Intelligence},
  volume={35},
  number={8},
  pages={7474--7482},
  year={2021}
}

@article{luo2024self,
  title={Self-Improved Learning for Scalable Neural Combinatorial Optimization},
  author={Luo, Fu and Lin, Xi and Wang, Zhenkun and Tong, Xialiang and Yuan, Mingxuan and Zhang, Qingfu},
  journal={arXiv preprint arXiv:2403.19561},
  year={2024}
}

@article{zhou2024instance,
  title={Instance-Conditioned Adaptation for Large-Scale Generalization of Neural Routing Solver},
  author={Zhou, Changliang and Lin, Xi and Wang, Zhenkun and Tong, Xialiang and Yuan, Mingxuan and Zhang, Qingfu},
  journal={IEEE Transactions on Intelligent Transportation Systems},
  volume={27},
  number={7},
  pages={8270--8284},
  year={2026},
  doi={10.1109/TITS.2026.3674538}
}

@article{croes1958method,
  title={A Method for Solving Traveling-Salesman Problems},
  author={Croes, Georges A.},
  journal={Operations Research},
  volume={6},
  number={6},
  pages={791--812},
  year={1958}
}

@article{lin1973effective,
  title={An Effective Heuristic Algorithm for the Traveling-Salesman Problem},
  author={Lin, Shen and Kernighan, Brian W.},
  journal={Operations Research},
  volume={21},
  number={2},
  pages={498--516},
  year={1973}
}

@techreport{helsgaun2017extension,
  title={An Extension of the {Lin-Kernighan-Helsgaun} {TSP} Solver for Constrained Traveling Salesman and Vehicle Routing Problems},
  author={Helsgaun, Keld},
  institution={Roskilde University},
  year={2017}
}

@article{vidal2022hybrid,
  title={Hybrid Genetic Search for the {CVRP}: Open-Source Implementation and {SWAP*} Neighborhood},
  author={Vidal, Thibaut},
  journal={Computers \& Operations Research},
  volume={140},
  pages={105643},
  year={2022}
}

@article{uchoa2017new,
  title={New Benchmark Instances for the Capacitated Vehicle Routing Problem},
  author={Uchoa, Eduardo and Pecin, Diego and Pessoa, Artur and Poggi, Marcus and Vidal, Thibaut and Subramanian, Anand},
  journal={European Journal of Operational Research},
  volume={257},
  number={3},
  pages={845--858},
  year={2017}
}

@inproceedings{drakulic2023bq,
  title={{BQ-NCO}: Bisimulation Quotienting for Efficient Neural Combinatorial Optimization},
  author={Drakulic, Darko and Michel, Sofia and Mai, Florian and Sors, Arnaud and Andreoli, Jean-Marc},
  booktitle={Advances in Neural Information Processing Systems},
  volume={36},
  year={2023}
}

@inproceedings{qiu2022dimes,
  title={{DIMES}: A Differentiable Meta Solver for Combinatorial Optimization Problems},
  author={Qiu, Ruizhong and Sun, Zhiqing and Yang, Yiming},
  booktitle={Advances in Neural Information Processing Systems},
  volume={35},
  year={2022}
}

@inproceedings{kim2022symnco,
  title={{Sym-NCO}: Leveraging Symmetricity for Neural Combinatorial Optimization},
  author={Kim, Minsu and Park, Junyoung and Park, Jinkyoo},
  booktitle={Advances in Neural Information Processing Systems},
  volume={35},
  year={2022}
}

@inproceedings{kwon2021matrix,
  title={Matrix Encoding Networks for Neural Combinatorial Optimization},
  author={Kwon, Yeong-Dae and Choo, Jinho and Yoon, Iljoo and Park, Minah and Park, Duwon and Gwon, Youngjune},
  booktitle={Advances in Neural Information Processing Systems},
  volume={34},
  year={2021}
}

@inproceedings{ma2022efficient,
  title={Efficient Neural Neighborhood Search for Pickup and Delivery Problems},
  author={Ma, Yining and Li, Jingwen and Cao, Zhiguang and Song, Wen and Guo, Hongliang and Gong, Yuejiao and Chee, Yeow Meng},
  booktitle={Proceedings of the Thirty-First International Joint Conference on Artificial Intelligence},
  pages={4776--4784},
  year={2022}
}

@inproceedings{xin2021neurolkh,
  title={{NeuroLKH}: Combining Deep Learning Model with {Lin-Kernighan-Helsgaun} Heuristic for Solving the Traveling Salesman Problem},
  author={Xin, Liang and Song, Wen and Cao, Zhiguang and Zhang, Jie},
  booktitle={Advances in Neural Information Processing Systems},
  volume={34},
  year={2021}
}

@inproceedings{hudson2022graph,
  title={Graph Neural Network Guided Local Search for the Traveling Salesperson Problem},
  author={Hudson, Benjamin and Li, Qingbiao and Malencia, Matthew and Prorok, Amanda},
  booktitle={International Conference on Learning Representations},
  year={2022}
}

@inproceedings{dwivedi2022graph,
  title={Graph Neural Networks with Learnable Structural and Positional Representations},
  author={Dwivedi, Vijay Prakash and Luu, Anh Tuan and Laurent, Thomas and Bengio, Yoshua and Bresson, Xavier},
  booktitle={International Conference on Learning Representations},
  year={2022}
}

@inproceedings{zhou2023towards,
  title={Towards Omni-generalizable Neural Methods for Vehicle Routing Problems},
  author={Zhou, Jianan and Wu, Yaoxin and Song, Wen and Cao, Zhiguang and Zhang, Jie},
  booktitle={Proceedings of the 40th International Conference on Machine Learning},
  series={Proceedings of Machine Learning Research},
  volume={202},
  pages={42769--42789},
  year={2023},
  publisher={PMLR}
}

@inproceedings{gensco,
  title={Generation as Search Operator for Test-Time Scaling of Diffusion-based Combinatorial Optimization},
  author={Li, Yang and Chen, Lvda and Wang, Haonan and Wang, Runzhong and Yan, Junchi},
  booktitle={Advances in Neural Information Processing Systems},
  volume={38},
  year={2025}
}

@inproceedings{maskco,
  title={{MaskCO}: Masked Generation Drives Effective Representation Learning and Exploiting for Combinatorial Optimization},
  author={Chen, Lvda and Li, Yang and Yan, Junchi},
  booktitle={International Conference on Learning Representations},
  year={2026}
}

@article{beardwood1959shortest,
  title={The Shortest Path through Many Points},
  author={Beardwood, Jillian and Halton, John H. and Hammersley, John M.},
  journal={Mathematical Proceedings of the Cambridge Philosophical Society},
  volume={55},
  number={4},
  pages={299--327},
  year={1959}
}

@misc{jordan2024muon,
  title={Muon: An Optimizer for Hidden Layers in Neural Networks},
  author={Jordan, Keller and Jin, Yuchen and Boza, Vlado and You, Jiacheng and Cesista, Franz and Newhouse, Laker and Bernstein, Jeremy},
  year={2024},
  howpublished={\url{https://kellerjordan.github.io/posts/muon/}}
}

@article{reinelt1991tsplib,
  title={{TSPLIB}---A Traveling Salesman Problem Library},
  author={Reinelt, Gerhard},
  journal={ORSA Journal on Computing},
  volume={3},
  number={4},
  pages={376--384},
  year={1991}
}

@article{clarke1964scheduling,
  title={Scheduling of Vehicles from a Central Depot to a Number of Delivery Points},
  author={Clarke, G. and Wright, J. W.},
  journal={Operations Research},
  volume={12},
  number={4},
  pages={568--581},
  year={1964}
}

\end{document}